\documentclass[journal]{IEEEtran}

\usepackage{amsmath,amssymb,amsfonts}
\usepackage{bm}
\usepackage{graphicx}
\usepackage{booktabs}
\usepackage{multirow}
\usepackage{makecell}
\usepackage{cite}
\usepackage[ruled,linesnumbered]{algorithm2e}
\usepackage{cite}
\usepackage[ruled,linesnumbered]{algorithm2e}
\usepackage[hidelinks]{hyperref}

\hypersetup{
  pdftitle={Beyond Control Points: Arcsecond Relative-Motion Estimation of Vision Measurement Platforms With Incomplete or Absent Control Fields},
  pdfauthor={Meng Lian, Jian Wang, Shuixin Pan, Haibo Liu, Yueqiang Zhang, and Yulan Guo}
}

\newtheorem{proposition}{Proposition}
\newtheorem{remark}{Remark}

\newcommand{\dt}{^{t_0,t_1}}
\newcommand{\skewm}[1]{\left[#1\right]_{\times}}

\graphicspath{{figs/}}

\begin{document}

\title{Beyond Control Points: Arcsecond Relative-Motion Estimation of Vision
Measurement Platforms With Incomplete or Absent Control Fields}

\author{Meng~Lian, Jian~Wang, Shuixin~Pan, Haibo~Liu,
Yueqiang~Zhang, and~Yulan~Guo%
\thanks{This manuscript has been submitted to \textit{IEEE Transactions on
Image Processing} for possible publication.}%
\thanks{This work was supported in part by the National Natural Science Foundation of China
under Grants 12372184 and 12002215, and in part by the Research Team Cultivation Program of Shenzhen University under Grant 2023JCT003.
(\textit{Corresponding author: Yueqiang Zhang}.)}%
\thanks{Meng Lian is with the Ministry of Natural Resources (MNR) Key
Laboratory for Geo-Environmental Monitoring of Great Bay Area \&
Guangdong Key Laboratory of Urban Informatics, Shenzhen University,
Shenzhen, China, and also with Shenzhen Expressway Co., Ltd., Shenzhen,
China.}%
\thanks{Jian Wang is with Shenzhen Expressway Co., Ltd., Shenzhen, China.}%
\thanks{Yueqiang Zhang and Shuixin Pan are with the Key Laboratory of
Optoelectronic Devices and Systems of Ministry of Education and
Guangdong Province, the Shenzhen Key Laboratory of Intelligent Optical
Measurement and Detection, and the College of Physics and
Optoelectronic Engineering, Shenzhen University, Shenzhen 518060,
China (e-mail: yueqiang.zhang@szu.edu.cn).}%
\thanks{Haibo Liu is with the School of Artificial Intelligence and Robotics,
Hunan University, Changsha 410082, China
(e-mail: haiboliu@hnu.edu.cn).}%
\thanks{Yulan Guo is with the School of Electronics and Communication Engineering, Sun Yat-sen University, Shenzhen 510275, China (e-mail: guoyulan@sysu.edu.cn).}}

\markboth{Preprint}%
{Beyond Control Points: Arcsecond Relative-Motion Estimation}

\maketitle

\begin{abstract}
Long-range vision-based deformation monitoring is highly sensitive to motion
of the camera platform. Absolute-pose differencing typically relies on
dedicated control data and propagates two independent pose errors into the
relative-motion estimate. We develop a control-adaptive differential
framework that estimates inter-frame platform motion directly from image
displacements and known 3D points. With no dedicated control point, the
framework recovers platform rotation from measurement-point observations.
One surveyed control point enables prior-constrained translation recovery,
while two nonparallel control rays recover full 3D translation. The framework
requires neither nonlinear optimization nor an initial pose estimate.
Excluding control data from the rotation stage makes the rotation estimate
exactly immune to contamination confined to the control field. The inherited
differential formulation also cancels translational extrinsic errors exactly.
We derive the rotation observability condition, a leakage bound for unmodeled
translation and nonrigid point motion, and the single-point axial-prior bias
law. Under 0.5-pixel image noise, attitude changes of up to 30~arcmin, and
3D point perturbations of up to 2~mm, the multi-camera estimator achieves a
rotation RMSE of 2.97~arcsec and an average runtime of 0.46~ms. With one
surveyed control point, its prior-constrained translation RMSE is 1.19~mm.
In a bridge experiment without a stable control field, the median
coordinate-wise displacement RMSE relative to total-station measurements
is 0.85~mm. The estimator also maintains zero divergence under the tested
3D coordinate perturbations on public RGB-D and stereo sequences. These
results establish state-of-the-art accuracy, calibration robustness, and
computational efficiency among the evaluated methods.
\end{abstract}


\section{Introduction}
\IEEEPARstart{A}{} telephoto camera monitoring a bridge from a distance of
300~m can resolve structural displacements below one millimeter. However,
even a small motion of the camera platform can produce an apparent
displacement comparable to or larger than the structural displacement of
interest. Deformation monitoring of large civil infrastructure
\cite{zhang2015,yu2015} therefore depends not only on image localization
accuracy but also on reliable compensation for platform motion. This study
addresses this motion-compensation problem.

Vision-based measurement techniques have attracted increasing attention
because of their noncontact operation, high spatial resolution, and ability
to support multipoint dynamic measurements. The image processing stage
typically includes optical flow and template registration
\cite{lucas1981,baker2004}, feature detection and tracking
\cite{shi1994,lowe2004,henriques2015}, digital image correlation for
full-field displacement and strain measurement~\cite{pan2009}, subpixel
phase correlation~\cite{foroosh2002,guizar2008}, and phase-based processing
for structural vibration measurement~\cite{wadhwa2013,chen2015jsv}.
These techniques are reviewed in~\cite{feng2018}.
These methods provide image-plane displacement observations. The remaining
challenge in field deployment is to convert pixel displacement into
structural displacement when the camera platform moves. Most existing
approaches assume a static platform. Environmental disturbances violate
this assumption, and even small platform motions produce substantial
displacement errors under long-range telephoto imaging. This study addresses
this estimation layer. Its inputs are the pixel-displacement observations
of control and measurement points, and its outputs are the platform motion
and the compensated structural displacement. The proposed formulation
permits translation recovery with one or two surveyed control points while
retaining the calibration-error behavior of the underlying differential
model.

Camera-platform 6-DOF pose estimation is a specialized instance of camera
pose estimation. When the camera intrinsics are known, the pose can be
estimated from correspondences between known 3D points and their 2D image
projections. This formulation is commonly known as the
Perspective-$n$-Point (PnP) problem. Its solution provides an absolute pose
estimate for each image.
Early research relied on optimization-based iterative algorithms. For instance, the LHM
method~\cite{lhm} minimizes reprojection error through orthogonal iteration,
while the gOp method~\cite{gop} casts the problem as a semidefinite program
(SDP) solved with off-the-shelf solvers. The PPnP method~\cite{ppnp} treats
it as an anisotropic orthogonal Procrustes problem. However, iterative
approaches are often computationally inefficient and prone to converging on
local rather than global optima.

To address the nonlinear complexity of iterative methods, researchers have
developed algorithms with linear complexity. The pioneering linear method,
EPnP~\cite{epnp}, utilizes four virtual control points to represent 3D points
and solves for their camera-space coordinates via null-space analysis. The
RPnP method~\cite{rpnp} adopts a ternary point set approach, establishing an
intermediate coordinate system based on the two most distant points to
simplify the rotation matrix. While these linear methods improve efficiency,
they do not guarantee global optimality.

Consequently, subsequent research has focused on achieving global optimality
alongside linear complexity, resulting in methods such as DLS~\cite{dls},
ASPnP~\cite{aspnp}, OPnP~\cite{opnp}, and optDLS~\cite{optdls}. DLS employs
the Cayley--Gibbs--Rodrigues (CGR) parameterization to simplify constraints
but suffers from singularities at 180-degree rotations. To circumvent this,
ASPnP utilizes quaternions and the Gr\"obner basis polynomial
method~\cite{cox}. Similarly, OPnP incorporates a point-set centering
strategy to enhance noise robustness. 
Recent developments include EOPnP~\cite{eopnp},
SQPnP~\cite{sqpnp}, and SRPnP~\cite{srpnp}. Despite these advances,
monocular PnP methods generally estimate optical-axis translation and roll
less accurately than pitch and yaw, particularly under a narrow field of
view. In vision-based deformation measurement, roll errors directly distort
the in-plane displacement compensation.

To improve the accuracy of 6-DOF pose estimation for platforms undergoing
small motions, Chen \emph{et al.}~\cite{chen2021} introduced a method based
on vertically aligned dual cameras. Although it improves accuracy relative
to monocular PnP, its fixed vertical configuration limits installation
flexibility. Yu \emph{et al.}~\cite{yu2022} and subsequent studies by the
same group~\cite{liu2024,yin2025,ge2025} developed generalized multi-camera
BPnP methods that permit flexible camera orientations. These methods,
however, require accurate initial values for nonlinear optimization and are
therefore relatively slow. In addition, they cannot recover the platform
pose when fewer than three control points are available.

When multiple cameras are rigidly mounted on the platform, pose estimation
can be formulated as a generalized noncentral PnP problem. In this
formulation, the observations are represented by spatial rays that do not
share a common optical center~\cite{pless}. Kneip \emph{et al.}
\cite{kneip2013} developed efficient solutions to the NPnP problem for
multi-camera robotic systems. The gOp method~\cite{gop} obtains globally
optimal estimates for general camera models through semidefinite
programming. The gDLS method~\cite{gdls} provides a scalable least-squares
solution to the generalized pose-and-scale problem, whereas UPnP
\cite{upnp} provides an optimal $O(n)$ solution applicable to both central
and noncentral cameras. Wientapper \emph{et al.}~\cite{wientapper} further
unified these absolute-pose formulations within a closed-form framework.

These generalized solvers independently estimate the absolute camera-rig
pose at each epoch. The relative platform motion is then obtained by
differencing the two estimates, which amplifies measurement noise. As shown
in Section~\ref{sec:sim}, their accuracy and efficiency are insufficient for
the arcsecond-level, high-frame-rate monitoring considered in this study.

Our previous work established the differential 6-DOF formulation, its
observability and calibration-error properties, and the validity boundaries
of the depth and first-order motion approximations
\cite{zhang2026differential}. The present study addresses a different
identifiability problem. In deformation monitoring, dedicated control points
may be incomplete or even absent, and the observations available for platform
rotation estimation may contain point-dependent structural displacement.

We therefore separate platform rotation from translation recovery. The
rotation stage fits the common platform-induced component of the
measurement-point observations, subject to an explicit rank condition and a
bound on the leakage caused by nonrigid point motion. When surveyed control
points are available, the translation stage uses one point together with an
optical-axis depth prior or uses two points with nonparallel viewing rays. The
resulting framework inherits the calibration-error properties of the
underlying differential model. It also remains operational below the
minimum dedicated-control-point requirement of absolute PnP methods.

The main contributions of this study are summarized as follows.
\begin{enumerate}
  \item Unlike our previous differential 6-DOF formulation, which establishes
  the general model and its calibration-error properties, this study
  determines what remains observable when the dedicated control field is
  incomplete or even absent. The resulting control-adaptive framework provides
  control-free rotation estimation, one-point prior-constrained translation,
  and two-point full 3D translation recovery.

  \item We establish the rank condition for rotation estimation from
  measurement-point observations and derive a leakage bound that quantifies
  the influence of unmodeled translation and nonrigid structural motion.
  For one-point translation recovery, we derive the exact
  $T/\sqrt{3}$ RMS axial-prior bias law and its applicability condition.

  \item We validate the framework through synthetic experiments, a bridge
  deformation experiment without a stable control field, and public RGB-D
  and stereo sequences. The multi-camera estimator achieves a rotation RMSE
  of $2.97''$, a one-point translation RMSE of 1.19~mm, and an average
  runtime of 0.46~ms, providing the best overall accuracy, robustness, and
  runtime among the evaluated methods.
\end{enumerate}

\section{Differential Measurement Model}
\label{sec:model}

\subsection{Camera Imaging Model}
The formulation applies to a monocular camera and to a rigid multi-camera
vision station. Let $W$, $B$, and $C$ denote the world, station, and camera
coordinate frames, respectively. A Euclidean 3D point expressed in frame
$A$ is denoted by $\bar{\bm{P}}^{A}\in\mathbb{R}^{3}$, and its homogeneous
form is $\bm{P}^{A}=[(\bar{\bm{P}}^{A})^{T}\;1]^{T}$. The homogeneous image
point is $\bm{p}^{t}=[u^{t},v^{t},1]^{T}$.
We define $\bm{G}_{A,B}\in SE(3)$ as the coordinate transformation from
frame $A$ to frame $B$, such that
$\bm{P}^{B}=\bm{G}_{A,B}\bm{P}^{A}$. The two-epoch projections are
\begin{equation}
\label{eq:proj}
\begin{aligned}
\lambda^{t_0}\bm{p}^{t_0} &=
  \left[\bm{K}\;\; \bm{0}_{3\times 1}\right]
  \bm{G}_{B,C}\,\bm{G}_{W,B_{t_0}}\bm{P}^{W} ,\\
\lambda^{t_1}\bm{p}^{t_1} &=
  \left[\bm{K}\;\; \bm{0}_{3\times 1}\right]
  \bm{G}_{B,C}\,\bm{G}_{B_{t_0},B_{t_1}}
  \bm{G}_{W,B_{t_0}}\bm{P}^{W} .
\end{aligned}
\end{equation}
where $\lambda^{t}$ is the camera-frame depth at epoch $t$,
$\bm{G}_{B,C}$ is the fixed camera extrinsic transformation, and
$\bm{G}_{B_{t_0},B_{t_1}}$ maps point coordinates from $B_{t_0}$ to
$B_{t_1}$. It is therefore the coordinate-change transformation induced by
the physical station motion. This convention determines the signs of the
estimated rotation and translation parameters. The same projection model is
applied independently to every camera in a rigid multi-camera station.

\subsection{Differential Model}
Following the differential 3D-2D formulation established in our previous
work~\cite{zhang2026differential}, we estimate the inter-epoch pose change
directly rather than differencing two independently estimated absolute poses.
In long-range monitoring, the depth change induced by small platform motion
may be small relative to the working distance. Introducing the approximation
$\lambda^{t_1}\approx\lambda^{t_0}$ and subtracting the two projection
equations in \eqref{eq:proj} gives
\begin{equation}
\label{eq:diff}
\Delta\bm{p}\dt=\frac{1}{\lambda^{t_0}}
  \left[\bm{K}\;\; \bm{0}_{3\times 1}\right]\bm{G}_{B,C}
  \left(\bm{G}_{B_{t_0},B_{t_1}}-\bm{E}\right)\bm{P}^{B_{t_0}} ,
\end{equation}
where $\Delta\bm{p}\dt=\bm{p}^{t_1}-\bm{p}^{t_0}$,
$\bm{P}^{B_{t_0}}$ is the 3D point expressed in the station frame at $t_0$,
and $\bm{E}$ is the $4\times4$ identity matrix. The quantitative validity
condition for the depth approximation is given in
Section~\ref{sec:bounds}.

The inter-epoch motion is represented on $SE(3)$. For
$\bm{\eta}=[\bm{\eta}_T^{T}\;\;\bm{\eta}_R^{T}]^{T}\in\mathbb{R}^{6}$,
define the wedge matrix
\begin{equation}
\label{eq:exp}
\begin{aligned}
\bm{\eta}^{\wedge}
&=\begin{bmatrix}
[\bm{\eta}_R]_{\times} & \bm{\eta}_T\\
\bm{0}_{1\times3} & 0
\end{bmatrix}
=\sum_{j=0}^{5}\eta_j\bm{G}_j,\\
\bm{G}_{B_{t_0},B_{t_1}}
&=\operatorname{Exp}(\bm{\eta}^{\wedge})
\approx\bm{E}+\bm{\eta}^{\wedge} .
\end{aligned}
\end{equation}
where $\bm{G}_j$ are the generators of $\mathfrak{se}(3)$,
$[\cdot]_{\times}$ denotes the skew-symmetric matrix, and the final
approximation retains the first-order term of the matrix exponential.
Substituting
\eqref{eq:exp} into \eqref{eq:diff}, the relative measurement model can be
rewritten as
\begin{equation}
\label{eq:lin}
\Delta\bm{p}\dt=\frac{1}{\lambda^{t_0}}
  \left[\bm{K}\;\; \bm{0}_{3\times 1}\right]\bm{G}_{B,C}
  \sum_{j=0}^{5}\eta_j\bm{G}_j\,\bm{P}^{B_{t_0}} .
\end{equation}
Expanding the generator action in \eqref{eq:lin} yields
\begin{equation}
\label{eq:model}
\Delta\bm{p}\dt=\frac{1}{\lambda^{t_0}}\bm{K}\bm{R}_{B,C}
 \left(-\skewm{\bar{\bm{P}}^{B_{t_0}}}\bm{\eta}_R+\bm{\eta}_T\right),
\end{equation}
where $\bm{R}_{B,C}$ is the rotational component of the camera extrinsics,
and $\skewm{\cdot}$ acts on a Euclidean 3D point.

For statistical estimation, we use the 2D pixel coordinates
$\bm{q}_i=[u_i,v_i]^{T}$ and the observation
$\Delta\bm{q}_i=\bm{q}_i^{t_1}-\bm{q}_i^{t_0}$. Let
$\tilde{u}_i=u_i^{t_0}-c_x$ and
$\tilde{v}_i=v_i^{t_0}-c_y$. The image Jacobian is
\begin{equation}
\label{eq:imagejac}
\bm{\Pi}_i=\frac{1}{\lambda_i^{t_0}}
\begin{bmatrix}
f_x & 0 & -\tilde{u}_i\\
0 & f_y & -\tilde{v}_i
\end{bmatrix}.
\end{equation}
If measurement point $i$ undergoes a structural displacement
$\Delta\bar{\bm{P}}_i$, its first-order pixel-difference model is
\begin{equation}
\label{eq:rotonly}
\begin{aligned}
\Delta\bm{q}_i
&=\bm{A}_i\bm{\eta}_R+\bm{B}_i\bm{\eta}_T
 +\bm{D}_i\Delta\bar{\bm{P}}_i+\bm{\epsilon}_i,\\
\bm{A}_i
&=-\bm{\Pi}_i\bm{R}_{B,C}
  \skewm{\bar{\bm{P}}_i^{B_{t_0}}},\qquad
\bm{B}_i=\bm{D}_i=\bm{\Pi}_i\bm{R}_{B,C}.
\end{aligned}
\end{equation}

The control-independent rotation stage fits the common rotational component
of the measurement-point observations. It does not assume that the
point-dependent structural motion vanishes. Instead, the unmodeled platform
translation and structural displacement are retained as explicit leakage
terms. Stacking the measurement points gives
\begin{equation}
\label{eq:stack}
\begin{aligned}
\bm{b}&=\bm{A}\bm{\eta}_R+\bm{r},\\
\widehat{\bm{\eta}}_R
&=\left(\bm{A}^{T}\bm{W}\bm{A}\right)^{-1}
  \bm{A}^{T}\bm{W}\bm{b},
\end{aligned}
\end{equation}
where $\bm{b}=\operatorname{col}_i(\Delta\bm{q}_i)$,
$\bm{W}$ is a positive-definite weight matrix, and
$\bm{r}=\operatorname{col}_i(\bm{B}_i\bm{\eta}_T
+\bm{D}_i\Delta\bar{\bm{P}}_i+\bm{\epsilon}_i)$. The rotation is locally
observable if and only if $\operatorname{rank}(\bm{A})=3$. Its deterministic
leakage error satisfies
\begin{equation}
\label{eq:leakage}
\begin{aligned}
\widehat{\bm{\eta}}_R-\bm{\eta}_R
&=\left(\bm{A}^{T}\bm{W}\bm{A}\right)^{-1}
  \bm{A}^{T}\bm{W}\bm{r},\\
\left\|\widehat{\bm{\eta}}_R-\bm{\eta}_R\right\|_2
&\leq
\frac{\|\bm{W}^{1/2}\bm{r}\|_2}
{\sigma_{\min}(\bm{W}^{1/2}\bm{A})}.
\end{aligned}
\end{equation}
Thus, nonrigid point motion is not automatically separable from platform
rotation. Accurate recovery requires a well-conditioned $\bm{A}$ and a small
projection of the omitted terms onto its rotation subspace. Independent
zero-mean point motions tend to average statistically as the number and
geometric diversity of the points increase, whereas coherent motion is
governed directly by \eqref{eq:leakage}.

\subsection{Translation and Structural-Displacement Recovery}
After rotation estimation, the rotation-compensated residual of a surveyed
control point $c$ is
\begin{equation}
\label{eq:transresid}
\bm{r}_c=\Delta\bm{q}_c-\bm{A}_c\widehat{\bm{\eta}}_R
=\bm{B}_c\bm{\eta}_T+\bm{\epsilon}_c.
\end{equation}
The matrix $\bm{B}_c\in\mathbb{R}^{2\times3}$ has rank two. Let
\begin{equation}
\label{eq:viewray}
\bm{d}_c=
\frac{\bm{R}_{B,C}^{T}\bm{K}^{-1}\bm{p}_c^{t_0}}
{\|\bm{R}_{B,C}^{T}\bm{K}^{-1}\bm{p}_c^{t_0}\|_2},
\qquad
\bm{n}_c=\bm{R}_{B,C}^{T}\bm{e}_3,
\end{equation}
where $\bm{d}_c$ is the control-point viewing ray and $\bm{n}_c$ is the
camera optical axis, both expressed in the station frame. The null space of
$\bm{B}_c$ is spanned by $\bm{d}_c$. In the one-point mode, the
depth-invariance prior $\bm{n}_c^{T}\bm{\eta}_T=0$ supplies the missing
constraint. Provided that $\bm{n}_c^{T}\bm{d}_c\neq0$, the estimate is
obtained from
\begin{equation}
\label{eq:onecp}
\begin{bmatrix}
\bm{B}_c^{T}\bm{W}_c\bm{B}_c & \bm{n}_c\\
\bm{n}_c^{T} & 0
\end{bmatrix}
\begin{bmatrix}
\widehat{\bm{\eta}}_T\\ \nu
\end{bmatrix}
=
\begin{bmatrix}
\bm{B}_c^{T}\bm{W}_c\bm{r}_c\\ 0
\end{bmatrix}.
\end{equation}
Thus, the image provides two independent translation constraints and the
third is supplied by the depth prior.

For two control points, possibly observed by different rigidly mounted
cameras, define
$\bm{B}_{12}=\operatorname{col}(\bm{B}_{c_1},\bm{B}_{c_2})$ and
$\bm{r}_{12}=\operatorname{col}(\bm{r}_{c_1},\bm{r}_{c_2})$. If the two
viewing rays are nonparallel, so that
$\operatorname{rank}(\bm{B}_{12})=3$, the complete 3D translation is
recovered without the depth prior:
\begin{equation}
\label{eq:twocp}
\widehat{\bm{\eta}}_T
=\left(\bm{B}_{12}^{T}\bm{W}_{12}\bm{B}_{12}\right)^{-1}
 \bm{B}_{12}^{T}\bm{W}_{12}\bm{r}_{12}.
\end{equation}

Once the observable platform motion has been determined, measurement point
$i$ satisfies
\begin{equation}
\label{eq:disp}
\begin{aligned}
\Delta\bm{q}_i
={}&\bm{\Pi}_i\bm{R}_{B,C}
\left[-\skewm{\bar{\bm{P}}_i^{B_{t_0}}}\bm{\eta}_R
+\bm{\eta}_T+\Delta\bar{\bm{P}}_i\right]\\
&+O\!\left(\|\bm{\eta}_R\|\,
\|\Delta\bar{\bm{P}}_i\|\right).
\end{aligned}
\end{equation}
After subtracting the platform-induced terms, the in-plane structural
displacement is recovered by weighted least squares under the adopted
constant-depth assumption. The depth component of a measurement point is
not observable from a single viewing ray.

The pixel-difference observation and the corresponding depth-scaled
normalized coordinates are algebraically equivalent only when the
observation covariance is transformed consistently. Uniform weighting in
the metric-scaled system changes the statistical model and assigns excessive
weight to distant targets. In the protocol of Section~\ref{sec:sim}, this
choice increases the translation RMSE by 34\% at a tenfold depth spread.
Lifting the differences to 3D before estimation introduces an artificial
zero-valued axial component and approximately doubles both rotation and
translation errors. We therefore formulate the estimator in pixel space,
which preserves the native image-noise metric and excludes the unobservable
axial component. Reproduction scripts are provided in the supplementary
code.

\subsection{Re-Linearization Pass}
\label{sec:relin}
To reduce the first-order truncation error at rotations of approximately
$\pm 30$~arcmin, we append one re-linearization pass to the initial linear
solution. Let $s\in\{0,1,2\}$ denote the number of surveyed control points
used for translation recovery. We write
$\mathcal{L}_s(\{\bm{P}_i^B\},\{\Delta\bm{q}_i\})$ for one application of
the corresponding deterministic linear estimator. It consists of
\eqref{eq:stack} and, when $s=1$ or $s=2$, the translation solve in
\eqref{eq:onecp} or \eqref{eq:twocp}, respectively. For $s=0$, it returns a
rotation-only increment with its translation fixed at zero.

Let $\widehat{\bm{\eta}}^{(1)}=\mathcal{L}_s(\{\bm{P}_i^B\},
\{\Delta\bm{q}_i\})$ and
$\widehat{\bm{G}}^{(1)}=\operatorname{Exp}
\big((\widehat{\bm{\eta}}^{(1)})^{\wedge}\big)$. The image displacement predicted by this
estimate is evaluated using the exact perspective projection. The residual is
\begin{equation}
\label{eq:resid}
\delta\bm{q}
=\Delta\bm{q}-\Delta\bm{q}\big(\widehat{\bm{G}}^{(1)}\big),
\end{equation}
where $\Delta\bm{q}(\widehat{\bm{G}}^{(1)})$ denotes the exact-projection
prediction. For a left-increment update, the reference points are transformed
by the first-pass estimate and the selected linear estimator is applied once
more to the residual:
\begin{equation}
\label{eq:secondpass}
\begin{aligned}
\bm{P}_i'{}^{B}
&=\widehat{\bm{G}}^{(1)}\bm{P}_i^{B},\\
\widehat{\bm{\eta}}^{(2)}
&=\mathcal{L}_s\big(\{\bm{P}_i'{}^{B}\},\{\delta\bm{q}_i\}\big),\\
\widehat{\bm{G}}
&=\operatorname{Exp}\big((\widehat{\bm{\eta}}^{(2)})^{\wedge}\big)
  \widehat{\bm{G}}^{(1)},\\
\widehat{\bm{\eta}}
&=\operatorname{Log}(\widehat{\bm{G}})^{\vee}.
\end{aligned}
\end{equation}
The same control mode is used in both passes. Thus, translation remains
unobserved in the zero-control mode, is prior-constrained in the one-control
mode, and is fully recovered when two nonparallel control rays are available.
The second pass reduces the leading truncation residual over the tested
motion range without introducing an iteration stopping rule. A
right-increment implementation would reverse the order of the two group
factors in \eqref{eq:secondpass}.

\begin{algorithm}[t]
\caption{Differential motion correction and displacement compensation}
\label{alg:pipeline}
\KwIn{two-epoch images $I(t_0)$ and $I(t_1)$; measurement points
$\{\bar{\bm{P}}^{B}_i\}$; zero, one, or two surveyed control points;
intrinsics $\bm{K}_k$, full rigid extrinsics $\bm{G}^{k}_{B,C}$, and
reference depths $\lambda_i$.}
\KwOut{platform rotation $\widehat{\bm{\eta}}_R$; platform translation
$\widehat{\bm{\eta}}_T$ when observable or prior-constrained; in-plane
structural displacements $\{\Delta\bar{\bm{P}}_j\}$.}
Extract and match the targets and form
$\Delta\bm{q}_i=\bm{q}_i^{t_1}-\bm{q}_i^{t_0}$\;
Construct $\bm{A}$ from the measurement points and verify
$\operatorname{rank}(\bm{A})=3$\;
Solve $\widehat{\bm{\eta}}_R^{(1)}$ using \eqref{eq:stack}\;
\eIf{one surveyed control point is available}{
  solve the prior-constrained translation using \eqref{eq:onecp}\;
}{
  \eIf{two surveyed control points are available and
  $\operatorname{rank}(\bm{B}_{12})=3$}{
    solve the complete 3D translation using \eqref{eq:twocp}\;
  }{
    report the rotation-only solution and leave translation unobserved\;
  }
}
Evaluate the exact-projection residual using \eqref{eq:resid}, apply one
left-increment update in the same control mode, and compose the estimates
using \eqref{eq:secondpass}\;
Recover each observable in-plane structural displacement using
\eqref{eq:disp}\;
\end{algorithm}

\begin{remark}
The rotation and two-point translation normal matrices are $3\times3$, and
the one-point constrained system is $4\times4$. Matrix assembly has
complexity $O(n)$. Omitting the residual update gives the single-pass
variant. No absolute pose is estimated at any stage.
\end{remark}

\section{Error-Bound Analysis}
\label{sec:bounds}
The differential formulation relies on several modeling approximations.
This section quantifies the image-space residuals introduced by the depth
approximation and the first-order motion model. It also analyzes the
influence of control-field contamination and the bias introduced by
depth-prior translation recovery. The depth and truncation analyses
specialize the validity boundaries established in our previous
work~\cite{zhang2026differential} to the present model with point-dependent
structural displacement. The leakage and low-control-point translation
analyses address the additional identifiability problem considered here.
The resulting conditions and error laws
are expressed in terms of the image-noise standard deviation $\sigma$ in
pixels, the focal length $f$ in pixels, the point depth $\lambda$, and the
pixel eccentricity
$\rho=\|\bm{p}-\bm{c}\|$ measured from the principal point $\bm{c}$. The
maximum eccentricity within the field of view is denoted by
$\rho_{\max}$. The theoretical predictions are evaluated quantitatively in
Section~\ref{sec:sim}, with the translation-related results examined in
Section~\ref{sec:sim-tsweep}.

\subsection{Validity Domain of the Depth-Invariance Assumption}
The depth-invariance hypothesis
($\lambda^{t_1}\approx\lambda^{t_0}$) underlying \eqref{eq:diff} yields an
exact expression for the discarded image term. For point $i$,
\begin{equation}
\label{eq:depthresid}
\bm{\varepsilon}_{d,i}
=\frac{\Delta\lambda_i}{\lambda_i^{t_0}}
  \big(\bm{p}_i^{t_1}-\bm{c}\big),
\qquad
\|\bm{\varepsilon}_{d,i}\|_2
=\frac{|\Delta\lambda_i|}{\lambda_i^{t_0}}\rho_i .
\end{equation}
The residual is a radial image scaling about the principal point. A point at
the principal point is unaffected, whereas points near the field boundary
are the most sensitive. The complete first-order depth change contains the
platform translation, the rotation-induced term, and the structural
displacement:
\begin{equation}
\label{eq:depthdecomp}
\Delta\lambda_i
=\bm{e}_3^{T}\bm{R}_{B,C}
\left(
\bm{\eta}_T-\skewm{\bar{\bm{P}}_i^{B}}\bm{\eta}_R
+\Delta\bar{\bm{P}}_i
\right).
\end{equation}

\begin{proposition}[Depth-approximation applicability]
\label{prop:depth}
The image residual introduced by the depth approximation does not exceed the
adopted noise scale $\sigma$ if and only if
\begin{equation}
\label{eq:tzbound}
\max_i
\left(
\frac{|\Delta\lambda_i|}{\lambda_i^{t_0}}\rho_i
\right)
\leq\sigma .
\end{equation}
\end{proposition}

For points within radius $\rho_{\max}$ and a rotation magnitude $\theta$, a
conservative sufficient condition is
\begin{equation}
\label{eq:depthsufficient}
\begin{aligned}
&\max_i\left[
\left|\bm{e}_3^{T}\bm{R}_{B,C}\bm{\eta}_T\right|
+\lambda_i\frac{\rho_i}{f}\theta
\right.\\
&\left.\hspace{3em}
+\left|\bm{e}_3^{T}\bm{R}_{B,C}
\Delta\bar{\bm{P}}_i\right|
\right]
\leq
\min_i\left(\lambda_i\frac{\sigma}{\rho_i}\right).
\end{aligned}
\end{equation}
At $\lambda=100$~m, $f=100{,}000$ pixels,
$\rho_{\max}=2200$ pixels, and $\theta=30'$,
the first-order rotation-induced depth term is 19.2~mm and contributes
approximately 0.42 pixel at the field boundary. Therefore, the simpler
condition
$|T_Z|\leq\lambda\sigma/\rho_{\max}$, which gives 9~mm at the field
boundary and approximately 20~mm at mid-field, applies only when the
rotation-induced and structural contributions to $\Delta\lambda_i$ are
negligible.
\subsection{First-Order Truncation and Re-Linearization}

The approximation in \eqref{eq:exp} retains only the first-order term of the
exponential map. Under the paraxial pure-rotation approximation, a rotation
with magnitude $\theta$ produces a leading second-order image residual of
approximately $\tfrac{1}{2}f\theta^2$ pixels.

\begin{proposition}[Conditional paraxial truncation scales]
\label{prop:trunc}
\begin{equation}
\label{eq:thetabound}
\varepsilon_{1,\mathrm{parax}}
\simeq\frac{1}{2}f\theta^2,
\qquad
\theta\lesssim\sqrt{\frac{2\sigma}{f}},
\end{equation}
when the nominal single-pass residual is required to remain below $\sigma$.
If the update in \eqref{eq:secondpass} cancels the leading quadratic term,
the nominal residual after the second pass is
\begin{equation}
\label{eq:thirdorder}
\varepsilon_{2,\mathrm{parax}}
\simeq\frac{1}{6}f\theta^3.
\end{equation}
\end{proposition}

These expressions are leading Taylor terms rather than global image-space
bounds. Their accuracy also depends on the pixel eccentricity, rotation axis,
translation, and depth residual in \eqref{eq:depthresid}. For
$f=10^{5}$ pixels and $\sigma=0.2$ pixel, the nominal single-pass threshold
is approximately $6.9'$. At $\theta=30'$, the nominal cubic term is
0.011 pixel. Section~\ref{sec:sim} evaluates the two-pass estimator over the
tested motion and field-of-view range.

\subsection{Exact Exclusion of Control-Field Contamination}
\begin{proposition}[Control-field exclusion]
\label{prop:immunity}
Let $\bm{z}_c$ contain all observations and surveyed coordinates belonging
only to the dedicated control field. If \eqref{eq:stack} is constructed
exclusively from measurement-point observations, then
\begin{equation}
\label{eq:controlimmunity}
\frac{\partial\widehat{\bm{\eta}}_R}{\partial\bm{z}_c}=\bm{0}.
\end{equation}
Therefore, the control-independent rotation estimate is exactly immune to
arbitrary contamination confined to the excluded control data.
\end{proposition}

This architectural property does not imply immunity to platform translation
or structural motion of the measurement points. These quantities enter
$\bm{r}$ in \eqref{eq:stack}, and their influence is bounded by
\eqref{eq:leakage}. The property also does not apply to an all-point variant
that includes control points in the rotation stage.

The camera-extrinsic behavior follows the differential model established in
our previous work~\cite{zhang2026differential}. Translational
camera-to-platform extrinsic errors cancel exactly, whereas rotational
extrinsic errors produce a bilinear perturbation of order
$O(\|\bm{\mu}_R\|\,\|\bm{\eta}\|)$. Its amplification depends on the
conditioning of the stacked observation matrix. Section~\ref{sec:sim-extrinsic}
evaluates this sensitivity over the tested calibration-error range.

\subsection{Depth-Prior Bias Law of the Translation Recovery}
For one control point, \eqref{eq:onecp} sets the optical-axis translation
component to zero. Let
$\cos\alpha_c=\bm{n}_c^{T}\bm{d}_c$, where $\alpha_c$ is the angle between
the camera optical axis and the control-point viewing ray.

\begin{proposition}[$T/\sqrt{3}$ axial-prior bias law]
\label{prop:tsqrt3}
In the noise-free case, the error introduced by the one-point prior is
\begin{equation}
\label{eq:priorbias}
\bm{e}_{\mathrm{prior}}
=-\bm{d}_c
\frac{\bm{n}_c^{T}\bm{\eta}_T}
{\bm{n}_c^{T}\bm{d}_c}.
\end{equation}
Thus, the omitted optical-axis component has magnitude
$|\bm{n}_c^{T}\bm{\eta}_T|$. If
$\|\bm{\eta}_T\|_2=T$ and its direction is uniformly distributed over the
unit sphere, then
\begin{equation}
\label{eq:tsqrt3}
\operatorname{RMS}_{\mathrm{axial}}=\frac{T}{\sqrt{3}},
\qquad
\operatorname{RMS}_{\mathrm{prior}}
=\frac{T}{\sqrt{3}|\cos\alpha_c|}.
\end{equation}
If the noise-only translation error is zero-mean, independent of the
translation direction, and has RMS $\sigma_0$, the total translation RMS is
\begin{equation}
\label{eq:quadrature}
\operatorname{RMS}_t(T)
=\sqrt{\sigma_0^{2}
+\frac{T^{2}}{3\cos^{2}\alpha_c}}.
\end{equation}
For the paraxial control-point geometry used in the experiments,
$|\cos\alpha_c|\approx1$, and \eqref{eq:quadrature} reduces to the
$T/\sqrt{3}$ law.
\end{proposition}

\begin{proposition}[Prior-bias negligibility]
\label{prop:selection}
For a general translation distribution, the one-point prior bias remains
below the noise-only floor when
\begin{equation}
\label{eq:selection}
\frac{
\sqrt{\mathbb{E}[(\bm{n}_c^{T}\bm{\eta}_T)^2]}
}{|\bm{n}_c^{T}\bm{d}_c|}
\leq\sigma_0.
\end{equation}
Under an isotropic fixed-magnitude model, this condition becomes
$T\leq\sqrt{3}|\cos\alpha_c|\sigma_0$. It is a bias-negligibility condition
for the one-point estimator, not evidence that one point is more accurate
than a full-rank two-point recovery.
\end{proposition}

\section{Experiments With Synthetic Data}
\label{sec:sim}
This section evaluates the recovery of small vision-station rotations using
Monte Carlo simulations. We compare the proposed estimator with
representative monocular PnP methods and generalized multi-camera NPnP
solvers. The simulated configuration represents a long-range bridge
monitoring scenario. The image resolution is $3840\times2160$ pixels, and
the equivalent focal length is 100{,}000 pixels. Measurement points are
randomly generated at distances of 50--100~m from the cameras.

Because the 3D coordinates of the measurement points are surveyed using a
total station in practice, uniformly distributed perturbations bounded by
$\pm2$~mm are added to emulate survey errors. The proposed estimator uses
these observations as measurement points without requiring a dedicated
control field. The comparison methods treat the same observations as
control points, ensuring that all methods receive identical inputs. A
translational extrinsic calibration error bounded by $\pm5$~mm is also
introduced. The multi-camera configuration consists of two cameras with a
$60^{\circ}$ angle between their optical axes, a lateral separation of
0.6~m, and identical intrinsic parameters.

The three attitude angles vary within $\pm30$~arcmin, and the translation
components vary within $\pm1$~mm. The proposed rotation estimator does not
recover translation. Translation is therefore treated as an unmodeled
disturbance when evaluating the robustness of the rigid-body-constrained
formulation. The points are projected using the pinhole model, after which
Gaussian localization noise is added at several noise levels.

We conduct monocular and multi-camera comparisons. In the monocular group,
the proposed linear rotation estimator, denoted Proposed-PnP, is compared
with LHM~\cite{lhm}, EPnP+GN~\cite{epnp}, RPnP~\cite{rpnp},
DLS~\cite{dls}, ASPnP~\cite{aspnp}, and OPnP~\cite{opnp}. These methods use
all measurement points observed by Camera~1 to estimate the absolute pose at
each epoch and then difference the two estimates.

In the multi-camera group, the proposed two-pass linear estimator, denoted
Proposed-BPnP, is compared with gOp~\cite{gop}, gDLS~\cite{gdls},
UPnP~\cite{upnp}, GAPS~\cite{wientapper}, EA-GPnP~\cite{eagpnp}, and the
optimization-based binocular differential method BPnP. The generalized
solvers represent the observations from both cameras as spatial rays,
estimate the absolute rig pose at each epoch, and difference the resulting
poses. Each camera observes half of the measurement points.

Each configuration is evaluated in 100 trials, except for the
typical-condition experiment, which uses 200 trials. Accuracy is measured
using the RMSE of the three attitude angles and the mean and median total
rotation errors. A trial is classified as divergent when the rotation error
exceeds $1^{\circ}$. Divergent trials are excluded from the RMSE and mean,
and their percentage is reported separately.

\subsection{Influence of the Number of Measurement Points}
\begin{figure}[!tb]
\centering
\includegraphics[width=0.9\columnwidth]{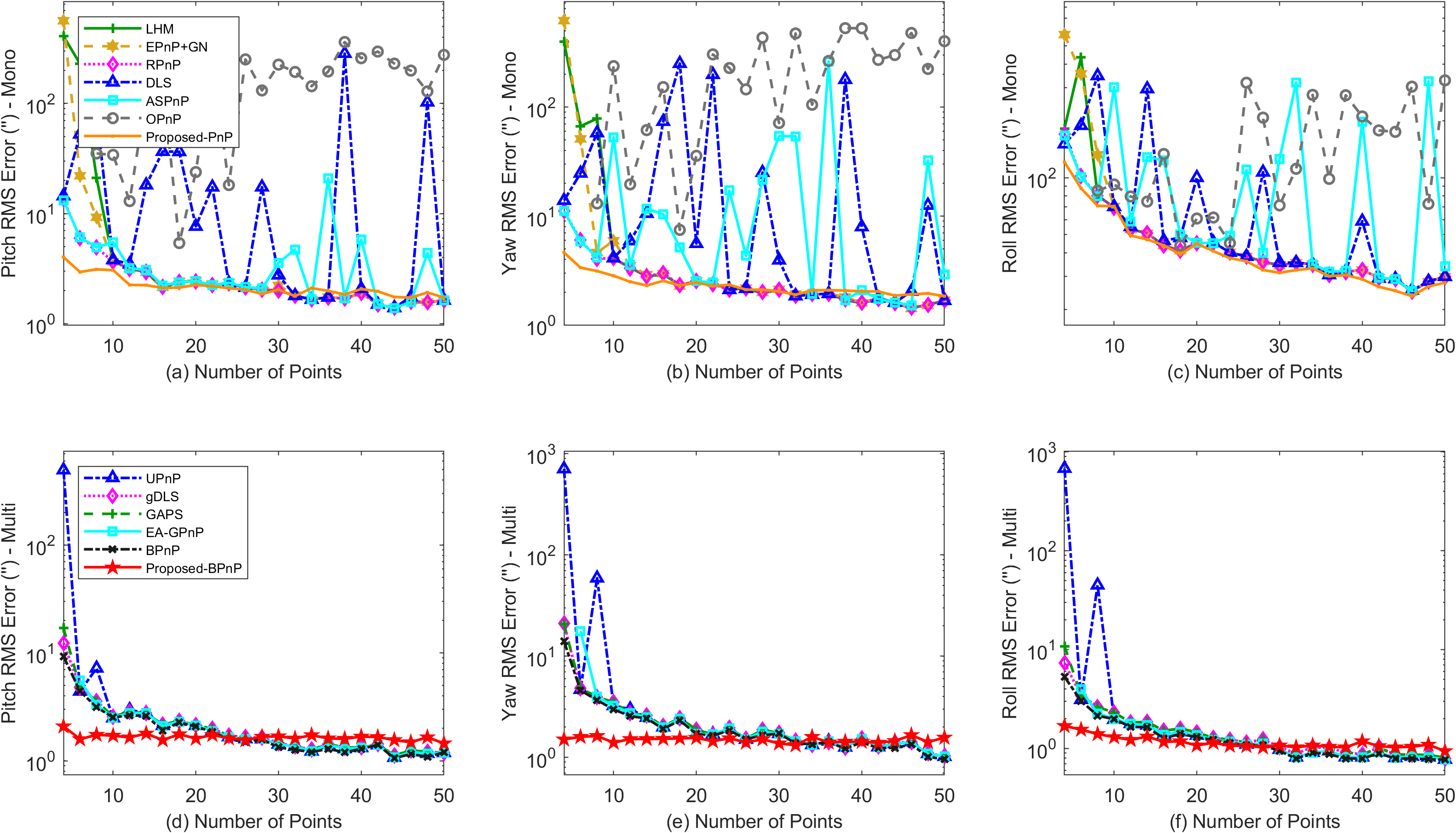}
\caption{RMSE of the three attitude angles versus the number of measurement
points: (a)--(c) monocular group; (d)--(f) multi-camera group.}
\label{fig:npts}
\end{figure}

The image noise is fixed at 0.5 pixel, representing the typical field
condition, while the number of measurement points varies from 4 to 50.
Figure~\ref{fig:npts} reports the RMSE of the three attitude angles on a
logarithmic vertical scale. The methods differ substantially when only a few
points are available. With four points, the pitch RMSE of LHM reaches
$410''$, and 19\% of its trials diverge. OPnP fails for fewer than eight
points and diverges sporadically at larger point counts. EA-GPnP fails in
the four-point case, whereas UPnP diverges in 33\% of the four-point trials.

With only two points per camera, Proposed-BPnP achieves RMSEs of
approximately $2.1''$, $1.5''$, and $1.7''$ for pitch, yaw, and roll,
respectively. Its RMSE remains close to $1.5''$ across the tested point
counts, with zero divergence. The two-pass solution reduces the
linearization error at rotations of up to $\pm30$~arcmin. The remaining
bias of approximately $1$--$1.5''$ is caused by the unmodeled
$\pm1$~mm translation and is largely independent of the number of points.

\subsection{Influence of Image Localization Error}
\begin{figure}[!tb]
\centering
\includegraphics[width=0.9\columnwidth]{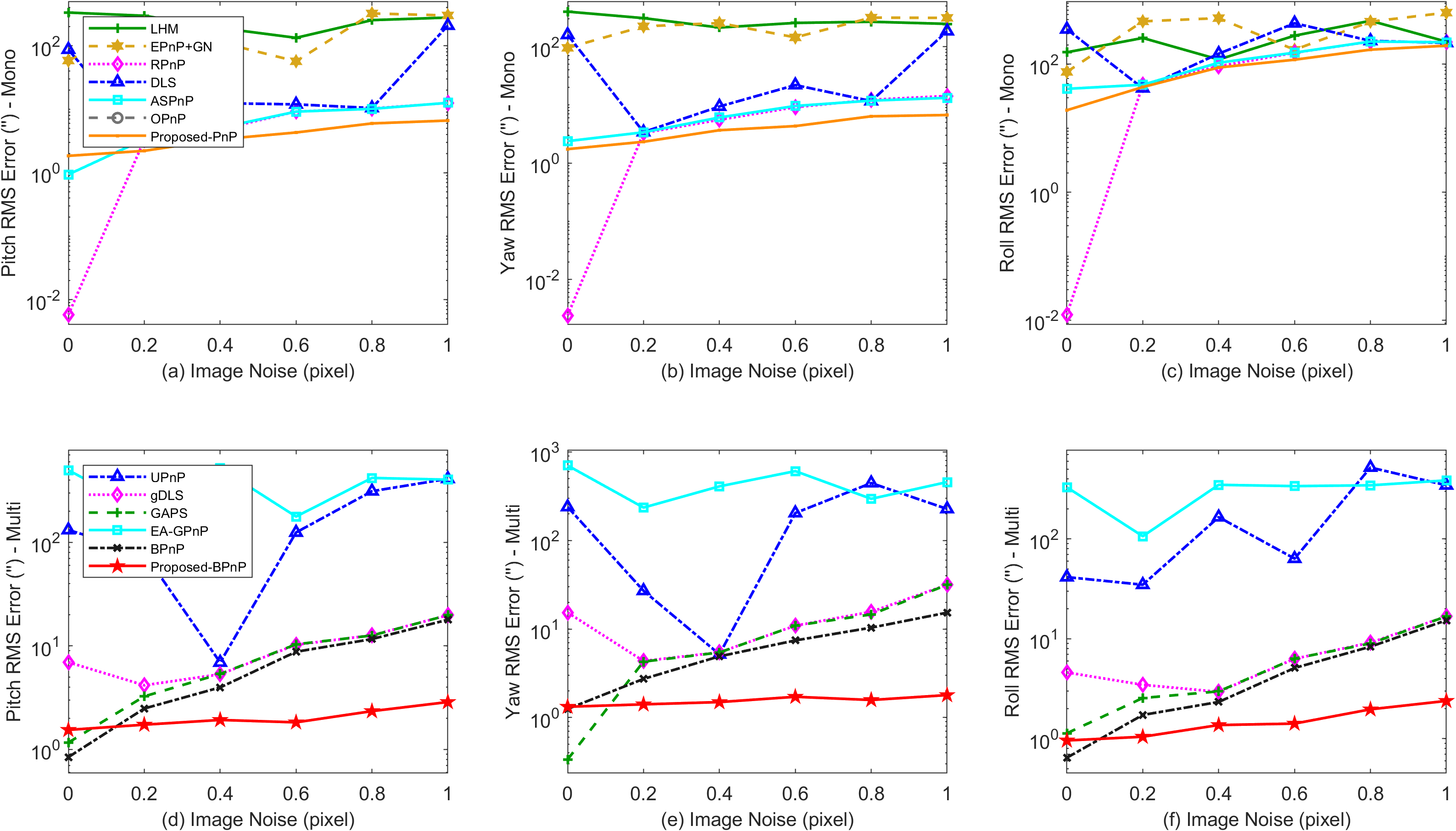}
\caption{RMSE of the three attitude angles versus the image localization
noise: (a)--(c) monocular group; (d)--(f) multi-camera group.}
\label{fig:noise}
\end{figure}

\begin{figure}[htbp]
\centering
\includegraphics[width=\columnwidth]{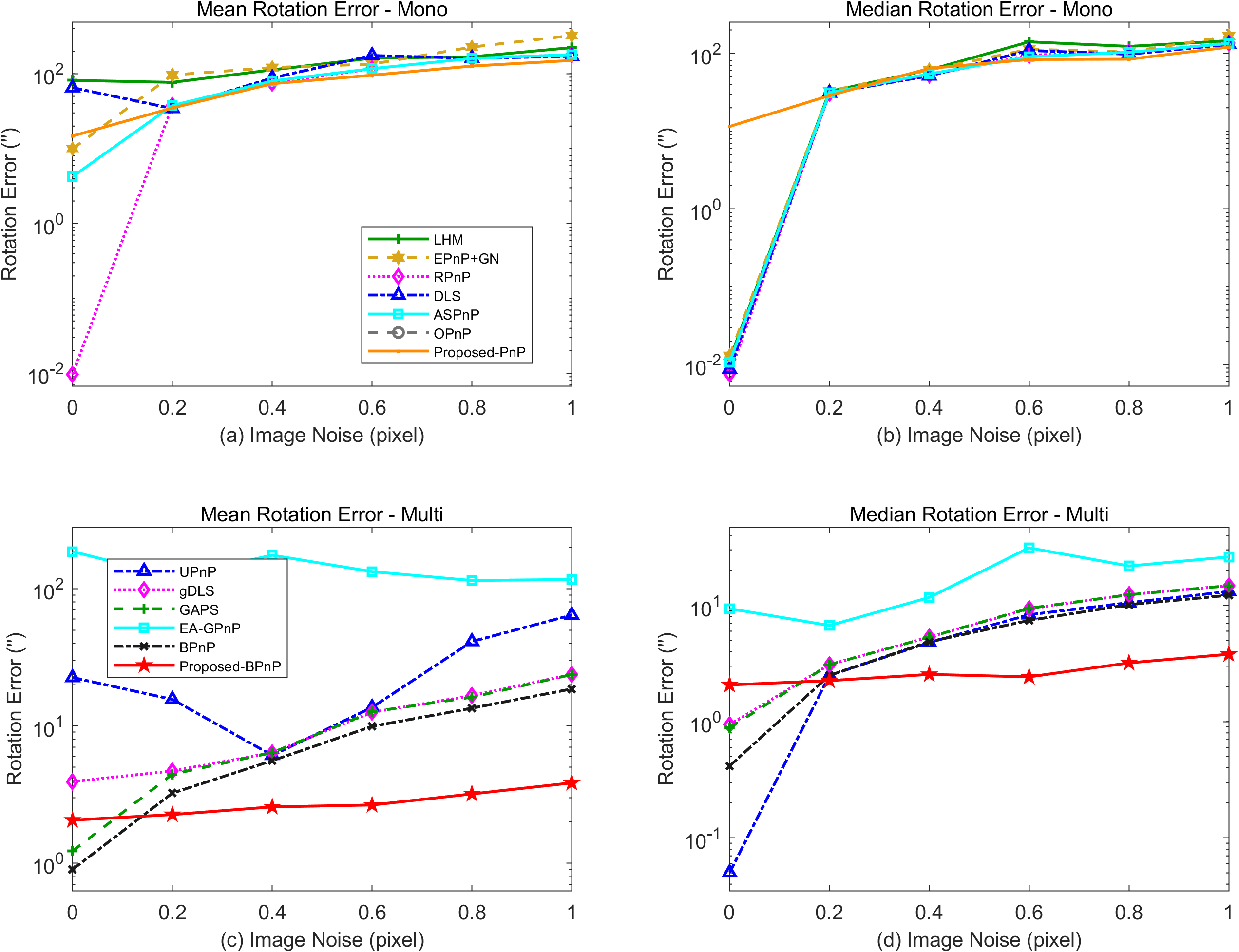}
\caption{Mean and median of the total rotation-angle error versus the image
localization noise: (a)(b) monocular group; (c)(d) multi-camera group.}
\label{fig:noisetot}
\end{figure}

With five measurement points, the image localization noise is varied from 0
to 1 pixel. Figure~\ref{fig:noise} reports the per-axis RMSE, and
Fig.~\ref{fig:noisetot} reports the mean and median total rotation errors.
At zero noise, the exact-model methods approach zero error. The proposed
estimators retain a bias of approximately $1.5''$ because translation of up
to $\pm1$~mm is not modeled by the rotation estimator.

The proposed methods achieve the lowest errors from a noise level of
0.2 pixel, and their advantage increases with the noise level. At 1-pixel
noise, the pitch, yaw, and roll RMSEs of Proposed-BPnP are $2.9''$,
$1.8''$, and $2.4''$, respectively. The corresponding BPnP errors are
$17.9''$, $15.3''$, and $15.3''$, while the gDLS and GAPS errors range from
$17''$ to $32''$. UPnP and EA-GPnP are affected by divergent trials.
Overall, the noise sensitivity of the proposed differential estimator is
approximately one sixth that of the absolute-pose differencing methods.

In the monocular comparison, LHM diverges in 6--17\% of the trials at every
noise level, and OPnP fails in the five-point configuration. Proposed-PnP
achieves the lowest pitch and yaw RMSEs without divergence. Above a noise
level of 0.4 pixel, its pitch and yaw RMSEs are approximately half those of
the strongest conventional method.

The separation between the mean and median curves in
Fig.~\ref{fig:noisetot} indicates heavy-tailed error distributions for LHM,
UPnP, and EA-GPnP. Their means exceed their medians by more than one order
of magnitude. In contrast, the mean and median errors of Proposed-BPnP
remain close, ranging from $1.5''$ to $2.9''$, which indicates a
substantially less heavy-tailed distribution under the tested conditions.

\subsection{Influence of Nonrigid Measurement-Point Motion}
\label{sec:sim-nonrigid}
We next add true inter-epoch structural displacements
$\Delta\bar{\bm{P}}_i$ independently of the survey-coordinate perturbations.
The monocular five-point configuration is evaluated using 0.5-pixel image
noise, rotations of up to $\pm30'$, platform translations of up to
$\pm1$~mm, survey perturbations of up to $\pm2$~mm, and 200 trials. Two
motion families are considered. The first contains independent zero-mean
in-plane point motions, whereas the second contains coherent motions aligned
with the least-observable rotation direction of $\bm{A}$.

As the pointwise displacement amplitude increases from 0.5 to 5~mm, the
rotation RMSE increases from $106''$ to $478''$ for the independent family
and from $178''$ to $1352''$ for the coherent family. The root-sum-square
combination of the deterministic prediction in \eqref{eq:leakage} and the
$98''$ noise-only floor agrees with the measured RMSE to within 1--7\%
throughout the sweep. The deterministic bound plus three noise floors covers
100\% of the independent trials and 99.9\% of the coherent trials. These
results confirm that nonrigid motion is not automatically removed by point
stacking and that the conditioning and residual alignment in
\eqref{eq:leakage} govern the rotation error. The supplementary script
\texttt{exp\_nonrigid\_leakage.m} reproduces this experiment.

\subsection{Influence of Measurement-Point Perturbation}
\begin{figure}[!tb]
\centering
\includegraphics[width=0.9\columnwidth]{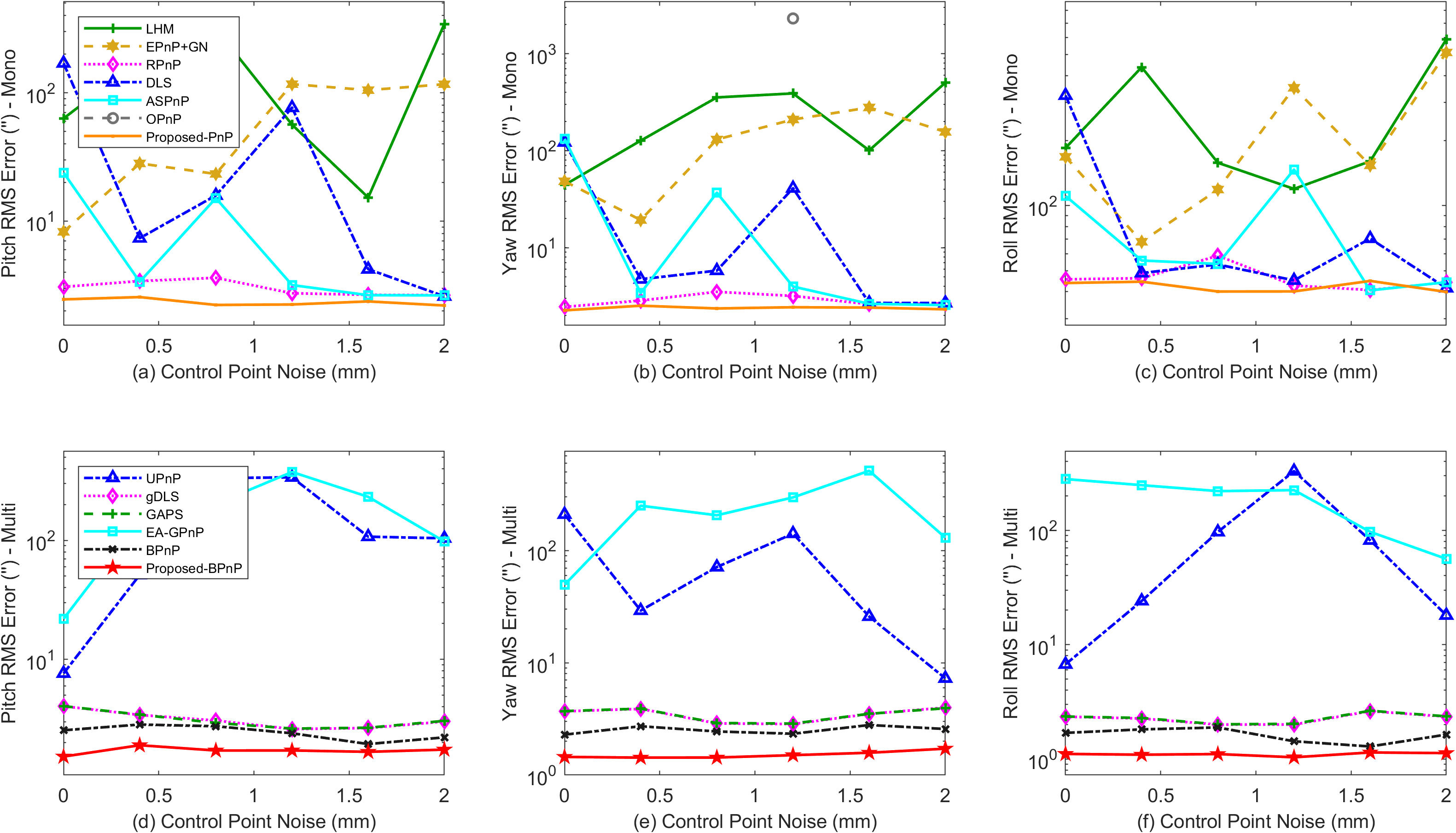}
\caption{RMSE of the three attitude angles versus the measurement-point
perturbation: (a)--(c) monocular group; (d)--(f) multi-camera group.}
\label{fig:ptperturb}
\end{figure}

Fig.~\ref{fig:ptperturb} shows the results when the measurement-point
perturbation grows from 0 to 2~mm with the image noise fixed at 0.2 pixel
and five measurement points. All the stable solutions are insensitive to the
perturbation: the three-axis RMSE of Proposed-BPnP stays within
$1.9''/1.7''/1.4''$, and gDLS/GAPS and BPnP remain flat as well, because the
influence of the measurement-point errors on the two instants is highly
correlated and largely cancels out in the differencing. Measurement points
surveyed with millimeter-level total-station accuracy are therefore
sufficient for arcsecond-level rotation measurement.

\subsection{Influence of Extrinsic Calibration Errors}
\label{sec:sim-extrinsic}
\begin{figure}[htbp]
\centering
\includegraphics[width=\columnwidth]{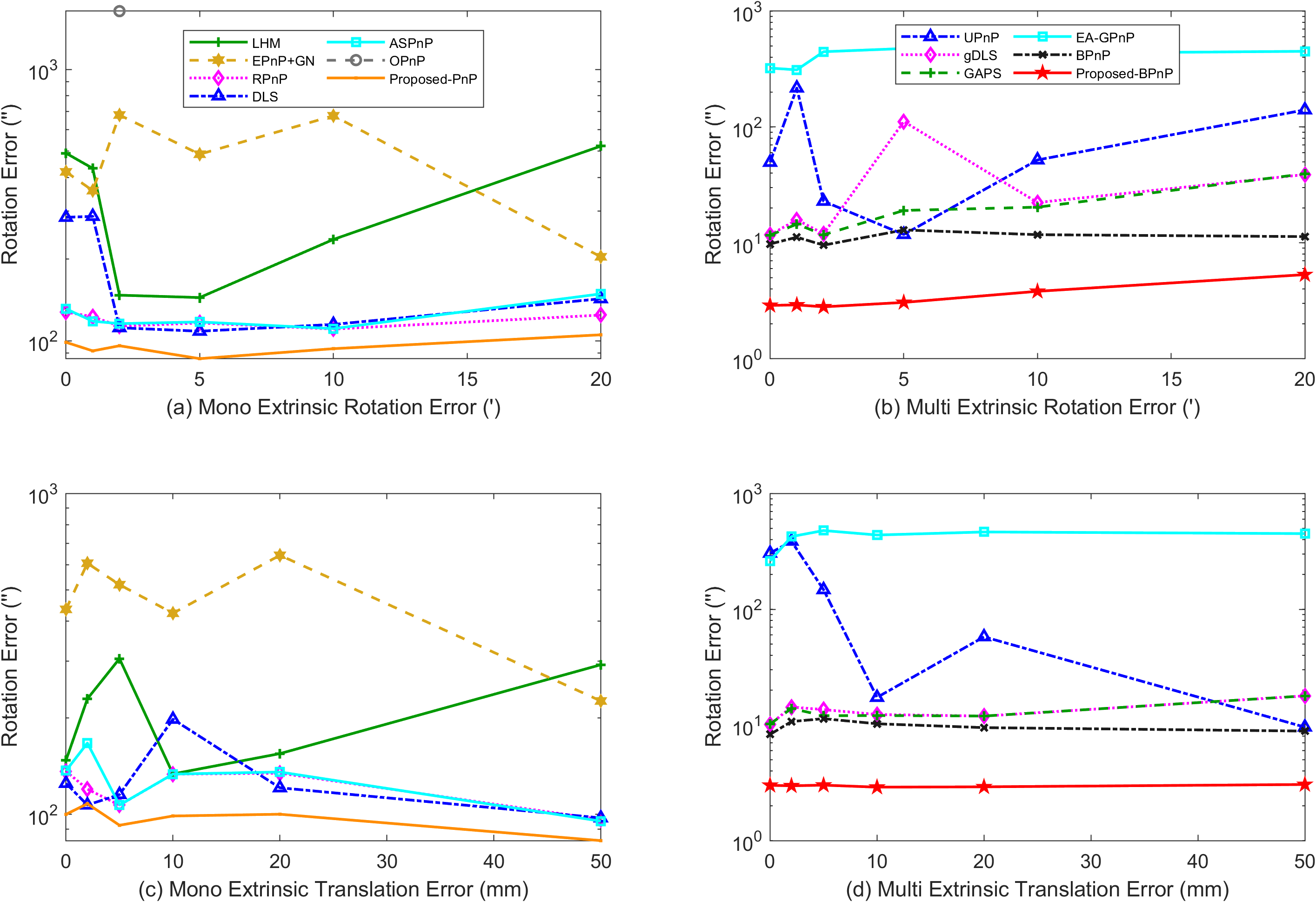}
\caption{Influence of the extrinsic calibration errors on the rotation
estimate: (a)(b) extrinsic rotation error; (c)(d) extrinsic translation
error.}
\label{fig:extrinsic}
\end{figure}

In practice the extrinsic parameters that relate the cameras to the platform
are obtained by total-station-assisted calibration and inevitably contain
errors. Fig.~\ref{fig:extrinsic} examines the influence of the extrinsic
rotation error (0 to 20 arcmin, with the translation error fixed at 5~mm)
and of the extrinsic translation error (0 to 50~mm, with no rotation error)
on the rotation estimate. Because the same erroneous extrinsics act at both
instants, their systematic effect largely cancels in the differencing
(Proposition~\ref{prop:immunity} and the remark thereafter): the rotation
RMSE of Proposed-BPnP is essentially insensitive to the extrinsic
translation error ($2.9$--$3.1''$ over the whole range) and grows only from
about $2.9''$ to $5.3''$ at the 20-arcmin extreme of the rotation error,
while Proposed-PnP stays at its noise-dominated level of $90$--$105''$.
Ordinary total-station-assisted calibration is therefore sufficient for the
proposed estimators.

\subsection{Influence of the Camera Mounting Angle}
\begin{figure}[htbp]
\centering
\includegraphics[width=0.9\columnwidth]{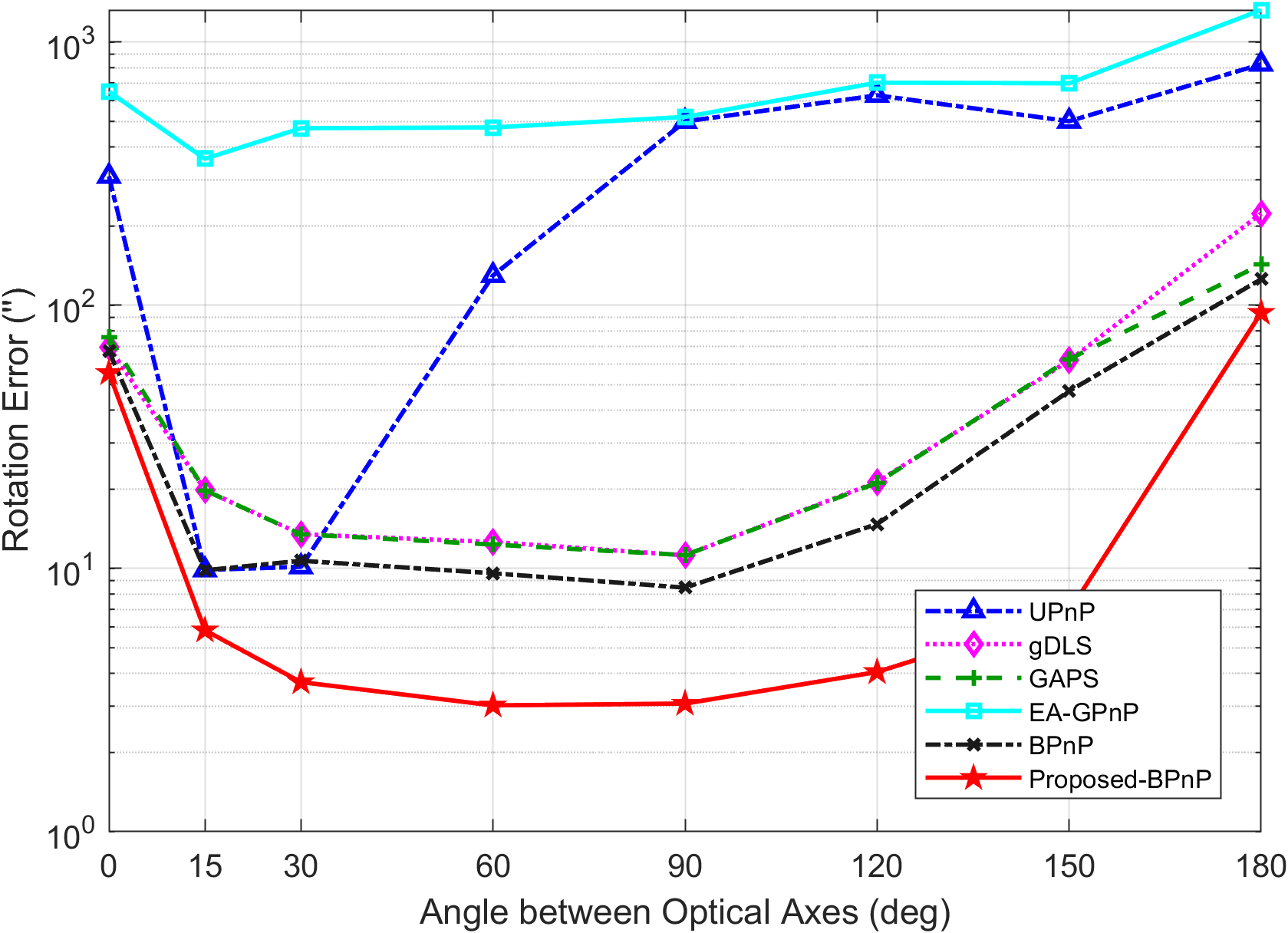}
\caption{Influence of the angle between the two optical axes on the rotation
estimate of the multi-camera group.}
\label{fig:mount}
\end{figure}

Fig.~\ref{fig:mount} shows the influence of the mounting angle between the
two optical axes, which varies from $0^{\circ}$ to $180^{\circ}$, on the
multi-camera group. The best accuracy is obtained for angles between
$30^{\circ}$ and $90^{\circ}$, where the rotation RMSE of Proposed-BPnP
stays at $3.0$--$3.7''$; when the two axes are nearly parallel
($0^{\circ}$) or opposite ($180^{\circ}$), the observation geometry
degenerates and the errors of all methods grow by more than an order of
magnitude. The $60^{\circ}$ angle adopted in the default configuration lies
in the optimal range.

\subsection{Statistical Distribution Under a Typical Condition}
\label{sec:sim-bound}
\begin{figure}[htbp]
\centering
\includegraphics[width=0.8\columnwidth]{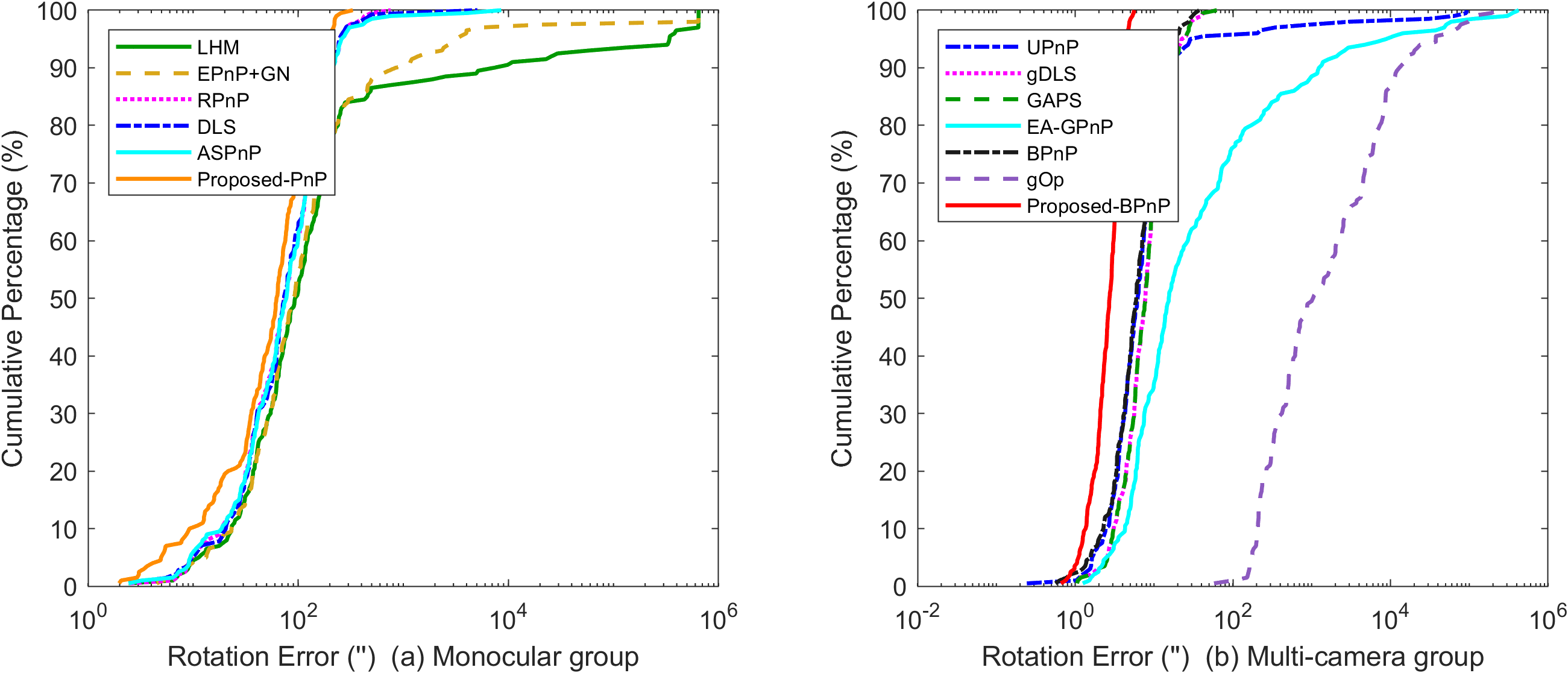}
\caption{Cumulative distribution of the total rotation-angle error under the
typical condition (0.5-pixel noise): (a) monocular group; (b) multi-camera
group.}
\label{fig:cdf}
\end{figure}

\begin{figure}[htbp]
\centering
\includegraphics[width=\columnwidth]{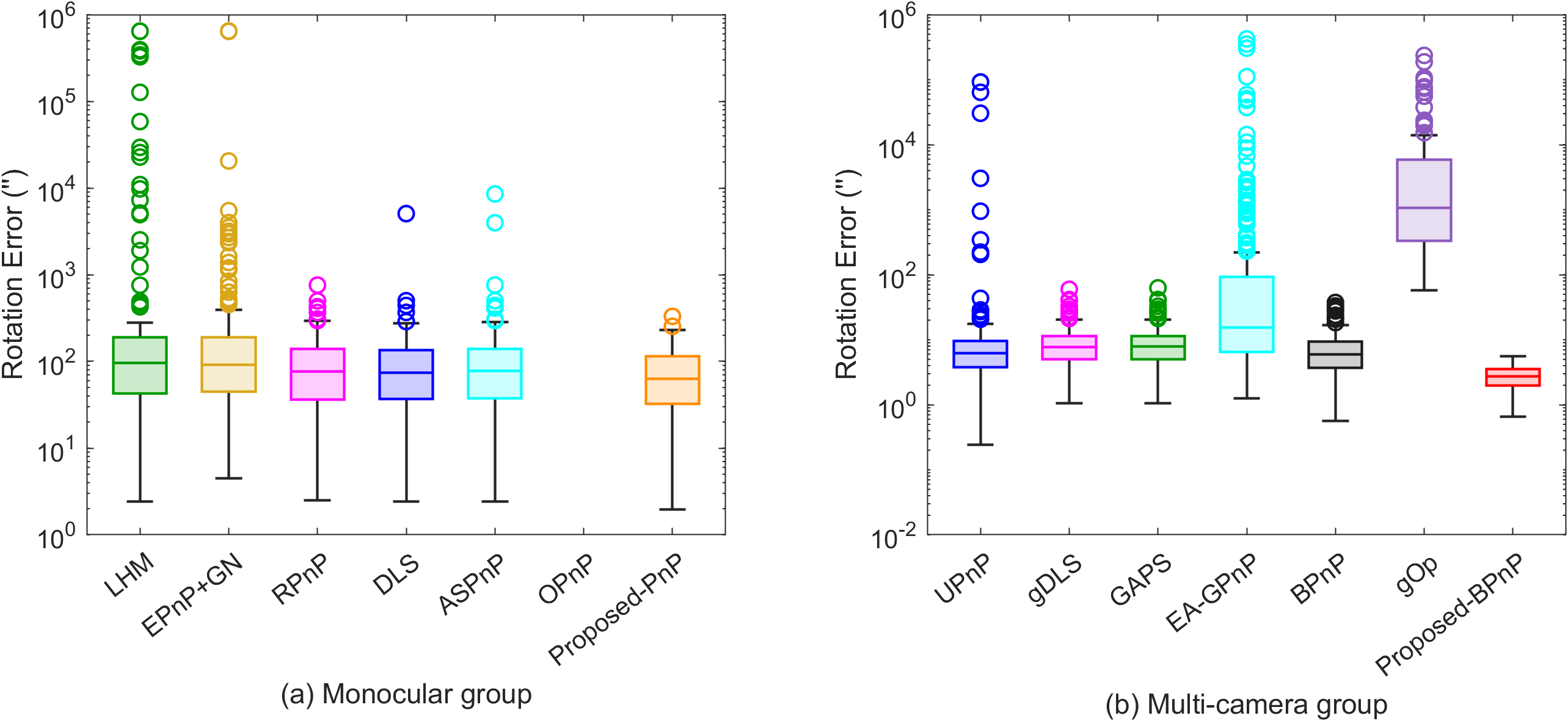}
\caption{Box plot of the rotation error under the typical condition:
(a) monocular group; (b) multi-camera group.}
\label{fig:box}
\end{figure}

We perform 200 Monte Carlo trials under a typical condition with five
measurement points, 0.5-pixel image noise, point perturbations of up to
$\pm2$~mm, rotations of up to $\pm30$~arcmin, and translations of up to
$\pm1$~mm. The 0.5-pixel noise level is consistent with the reprojection
errors of 0.50--0.57 pixel measured in the field experiment of
Section~\ref{sec:real}. Figure~\ref{fig:cdf} presents the cumulative
distribution of the total rotation error, Fig.~\ref{fig:box} presents the
corresponding box plots, and Table~\ref{tab:typical} reports the numerical
statistics.

Since the roll component is barely observable in the monocular geometry and
its large errors would mask the differences in the two well-observed angles,
the monocular entries of Table~\ref{tab:typical} report the statistics of
the combined pitch--yaw error --- the components that govern the
virtual-displacement compensation at long working distances --- whereas the
multi-camera entries report the total rotation angle, since all three axes
are observable in that configuration; trials whose total rotation error
exceeds $1^{\circ}$ are counted as divergent in both groups. In the
monocular group, Proposed-PnP attains the best pitch--yaw RMSE ($5.3''$) and
median ($3.5''$) without any divergence, less than half of RPnP ($11.9''$)
and ASPnP ($12.0''$), while DLS diverges in 30\% of the trials, LHM in
11.5\%, and OPnP fails completely with five points. In the multi-camera
group, Proposed-BPnP achieves a total-rotation RMSE of $2.97''$ with a
maximum error of only $5.6''$ and zero divergence, about $3\times$ better
than BPnP ($9.42''$) and $4\times$ better than gDLS/GAPS ($12.0''$); UPnP
attains a lower median ($6.1''$) but diverges in 2\% of the trials with a
heavy sub-degree tail (filtered RMSE $230''$); EA-GPnP shows a moderate
median ($13.8''$) but a 6.5\% divergence rate, and the SDP-based gOp yields
errors of several hundred arcseconds with a 34\% divergence rate. The two
proposed estimators never diverged in any trial.

\begin{table}[t]
\centering
\caption{Rotation-Error Statistics (Arcsec) and Average Runtime Under the
Typical Condition (Pitch--Yaw Error for the Monocular Group; Total Rotation
Angle for the Multi-Camera Group)}
\label{tab:typical}
\setlength{\tabcolsep}{3.5pt}
\begin{tabular}{lcccccc}
\toprule
Method & \makecell{Mean\\$('')$} & \makecell{Median\\$('')$}
       & \makecell{RMSE\\$('')$} & \makecell{Max\\$('')$}
       & \makecell{Div.\\$(\%)$} & \makecell{Time\\(ms)}\\
\midrule
\multicolumn{7}{c}{\emph{Monocular methods (pitch--yaw error)}}\\
LHM         & 47.97 & 8.14 & 248.12 & 2524.57 & 11.5 & 5.01\\
EPnP+GN     & 40.02 & 8.47 & 138.93 & 1423.42 & 4.5  & 4.06\\
RPnP        & 9.06  & 7.05 & 11.94  & 55.65   & 0    & 2.42\\
DLS         & 10.38 & 7.49 & 17.16  & 141.29  & 30   & 5.62\\
ASPnP       & 9.07  & 7.00 & 12.03  & 55.15   & 1.0  & 5.39\\
OPnP        & ---   & ---  & ---    & ---     & 100  & ---\\
Proposed-PnP& \textbf{4.38} & \textbf{3.46} & \textbf{5.33} & \textbf{20.07} & \textbf{0} & \textbf{0.54}\\
\midrule
\multicolumn{7}{c}{\emph{Multi-camera methods (total rotation angle)}}\\
UPnP        & 31.27 & 6.07 & 229.85 & 3037.90 & 2.0  & 1.80\\
gDLS        & 9.41  & 7.79 & 12.00  & 59.69   & 0    & 9.67\\
GAPS        & 9.43  & 7.83 & 12.08  & 62.77   & 0    & 7.30\\
EA-GPnP     & 159.95& 13.85& 467.64 & 2909.00 & 6.5  & 7.09\\
BPnP        & 7.45  & 5.98 & 9.42   & 37.46   & 0    & 9.55\\
gOp         & 795.47& 489.56&1109.96& 3190.74 & 34.0 & 285.7\\
Proposed-BPnP&\textbf{2.77}&\textbf{2.74}&\textbf{2.97}&\textbf{5.63}&\textbf{0}&\textbf{0.46}\\
\bottomrule
\end{tabular}
\end{table}

\subsection{Translation Recovery With One or Two Control Points}
\label{sec:sim-hybrid}
The rotation-only mode requires no control points. When one or two surveyed
control points are available, the platform translation can also be
recovered. With the rotation fixed, the rotation-induced displacement of
each control point is removed using the exact projection. One control point
supplies two independent image constraints, while the optical-axis depth
prior in \eqref{eq:onecp} supplies the third constraint. Two nonparallel
control rays satisfy $\operatorname{rank}(\bm{B}_{12})=3$ and recover the
complete 3D translation without this prior. Section~\ref{sec:sim-tsweep}
evaluates the one-point prior-bias law.

For a fair comparison, the PnP and NPnP solvers use the five perturbed
measurement points together with the available control points to estimate
both rotation and translation. The all-point variant of the proposed
estimator uses the same point set for rotation and reserves the surveyed
control points for translation. Consequently, the exact control-field
exclusion property in Proposition~\ref{prop:immunity} does not apply to this
comparison variant.

Table~\ref{tab:onecp} reports the single-control-point results. The hybrid
scheme achieves a pitch--yaw RMSE of $4.50''$ and a translation RMSE of
1.21~mm in the monocular configuration. In the multi-camera configuration,
the corresponding errors are $2.86''$ and 1.19~mm. These errors are
approximately $2$--$2.5\times$ lower in rotation and $2\times$ lower in
translation than those of the strongest competitor, the iterative BPnP
method. The translation error is more than one order of magnitude lower than
those of the PnP solvers and gDLS/GAPS.

\begin{figure}[t]
\centering
\includegraphics[width=\columnwidth]{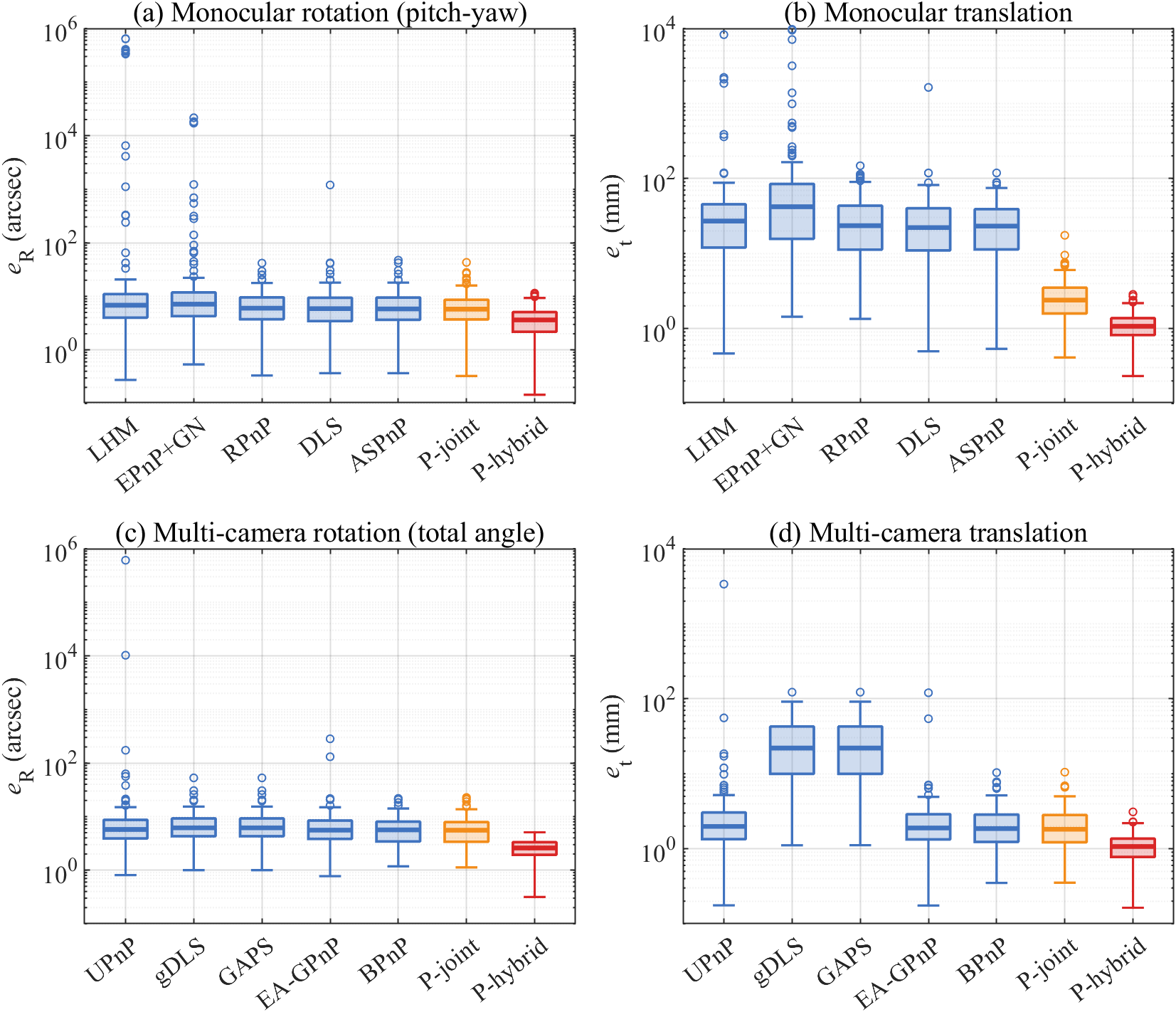}
\caption{Distributions of the rotation and translation errors under the
all-point protocol with one control point: (a) monocular rotation
(pitch--yaw); (b) monocular translation; (c) multi-camera rotation (total
angle); (d) multi-camera translation.}
\label{fig:cp1box}
\end{figure}

\begin{figure}[t]
\centering
\includegraphics[width=\columnwidth]{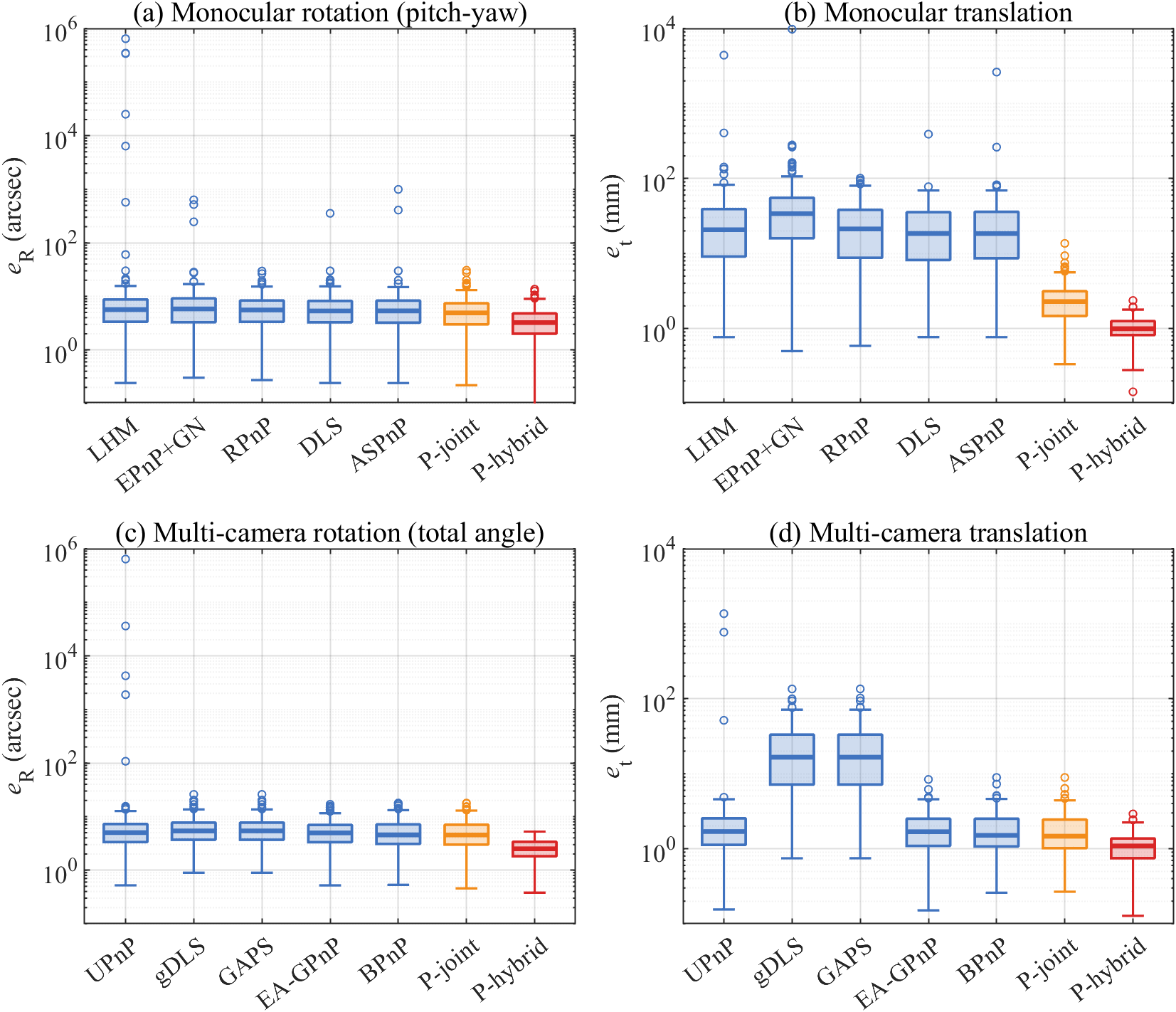}
\caption{Distributions of the rotation and translation errors under the
all-point protocol with two control points; panels as in
Fig.~\ref{fig:cp1box}.}
\label{fig:cp2box}
\end{figure}

Including the control point in the rotation stage also improves the all-point
variant. Under the same perturbed-point protocol, the monocular pitch--yaw
RMSE decreases from $10.1''$ with measurement points alone to $4.5''$ when
the control point is included. The joint 6-DOF solution using all points
remains $2$--$3\times$ less accurate in translation than the decoupled
scheme. Table~\ref{tab:twocp} reports the two-control-point case. The
multi-camera hybrid estimator achieves a rotation RMSE of $2.81''$ and a
translation RMSE of 1.21~mm without divergence.
Figs.~\ref{fig:cp1box} and~\ref{fig:cp2box} show that the hybrid estimator
has the most compact error distributions in both settings. A single stable
control point therefore supports millimeter-level, prior-constrained
translation together with arcsecond-level rotation, while two suitable
control rays remove the optical-axis prior.

\begin{table}[t]
\centering
\caption{Error Statistics Under the All-Point Protocol With One Surveyed
Control Point (Typical Condition, 200 Trials)}
\label{tab:onecp}
\scriptsize
\setlength{\tabcolsep}{2.2pt}
\begin{tabular}{lcccccc}
\toprule
Method & \makecell*[c]{Rot.\\RMSE/$''$} & \makecell*[c]{Rot.\\med./$''$}
       & \makecell*[c]{Trans.\\RMSE/mm} & \makecell*[c]{Trans.\\med./mm}
       & \makecell*[c]{$\Delta Z$\\RMSE/mm} & Div.\,\%\\
\midrule
\multicolumn{7}{c}{\emph{Monocular methods (pitch--yaw error)}}\\
LHM        & 89.67 & 6.40 & 34.64 & 27.07 & 34.48 & 5.0\\
EPnP+GN    & 52.30 & 6.93 & 44.99 & 42.02 & 44.82 & 3.0\\
RPnP       & 9.27  & 6.01 & 36.46 & 23.49 & 36.31 & 0\\
DLS        & 91.46 & 5.90 & 33.82 & 22.13 & 33.66 & 13.0\\
ASPnP      & 9.71  & 5.83 & 33.97 & 23.21 & 33.81 & 0\\
OPnP       & ---   & ---  & ---   & ---   & ---   & 100\\
Prop.-mono (joint)  & 8.64 & 5.78 & 3.34 & 2.39 & 2.16 & 0\\
Prop.-mono (hybrid) & \textbf{4.50} & \textbf{3.63} & \textbf{1.21} & \textbf{1.07} & \textbf{0.56} & \textbf{0}\\
\midrule
\multicolumn{7}{c}{\emph{Multi-camera methods (total rotation angle)}}\\
UPnP       & 15.76 & 5.67 & 5.13  & 1.99  & 1.14  & 1.0\\
gDLS       & 9.20  & 6.18 & 35.92 & 21.76 & 27.49 & 0\\
GAPS       & 9.20  & 6.18 & 35.94 & 21.99 & 27.50 & 0\\
EA-GPnP    & 23.18 & 5.60 & 4.59  & 1.89  & 4.62  & 0\\
BPnP       & 7.28  & 5.66 & 2.60  & 1.86  & 1.11  & 0\\
Prop.-bino (joint)  & 7.13 & 5.57 & 2.54 & 1.82 & 1.09 & 0\\
Prop.-bino (hybrid) & \textbf{2.86} & \textbf{2.60} & \textbf{1.19} & \textbf{1.07} & \textbf{0.57} & \textbf{0}\\
\bottomrule
\end{tabular}
\end{table}

\begin{table}[t]
\centering
\caption{Error Statistics Under the All-Point Protocol With Two Surveyed
Control Points (One per Camera in the Multi-Camera Configuration; Both
Viewed by the Single Camera in the Monocular Configuration)}
\label{tab:twocp}
\scriptsize
\setlength{\tabcolsep}{2.2pt}
\begin{tabular}{lcccccc}
\toprule
Method & \makecell*[c]{Rot.\\RMSE/$''$} & \makecell*[c]{Rot.\\med./$''$}
       & \makecell*[c]{Trans.\\RMSE/mm} & \makecell*[c]{Trans.\\med./mm}
       & \makecell*[c]{$\Delta Z$\\RMSE/mm} & Div.\,\%\\
\midrule
\multicolumn{7}{c}{\emph{Monocular methods (pitch--yaw error)}}\\
LHM        & 41.86 & 5.58 & 30.47 & 20.73 & 30.36 & 3.0\\
EPnP+GN    & 60.92 & 5.83 & 42.12 & 33.97 & 42.00 & 0\\
RPnP       & 7.78  & 5.57 & 33.09 & 21.26 & 32.96 & 0\\
DLS        & 27.08 & 5.33 & 28.75 & 18.47 & 28.61 & 6.5\\
ASPnP      & 7.54  & 5.31 & 29.95 & 18.43 & 29.82 & 1.0\\
OPnP       & 139.83& 5.63 & 32.35 & 21.64 & 32.22 & 2.5\\
Prop.-mono (joint)  & 7.23 & 4.91 & 3.03 & 2.29 & 2.08 & 0\\
Prop.-mono (hybrid) & \textbf{4.26} & \textbf{3.23} & \textbf{1.08} & \textbf{0.99} & \textbf{0.56} & \textbf{0}\\
\midrule
\multicolumn{7}{c}{\emph{Multi-camera methods (total rotation angle)}}\\
UPnP       & 134.76& 5.01 & 4.24  & 1.70  & 2.38  & 1.5\\
gDLS       & 7.09  & 5.39 & 29.44 & 16.59 & 22.95 & 0\\
GAPS       & 7.10  & 5.40 & 28.65 & 16.59 & 22.98 & 0\\
EA-GPnP    & 6.33  & 4.97 & 2.23  & 1.68  & 0.97  & 0\\
BPnP       & 6.34  & 4.55 & 2.23  & 1.51  & 0.96  & 0\\
Prop.-bino (joint)  & 6.21 & 4.55 & 2.18 & 1.47 & 0.94 & 0\\
Prop.-bino (hybrid) & \textbf{2.81} & \textbf{2.51} & \textbf{1.21} & \textbf{1.09} & \textbf{0.69} & \textbf{0}\\
\bottomrule
\end{tabular}
\end{table}

\subsection{Verification of the Error Bounds}
\label{sec:sim-tsweep}
\begin{figure}[htbp]
\centering
\includegraphics[width=0.9\columnwidth]{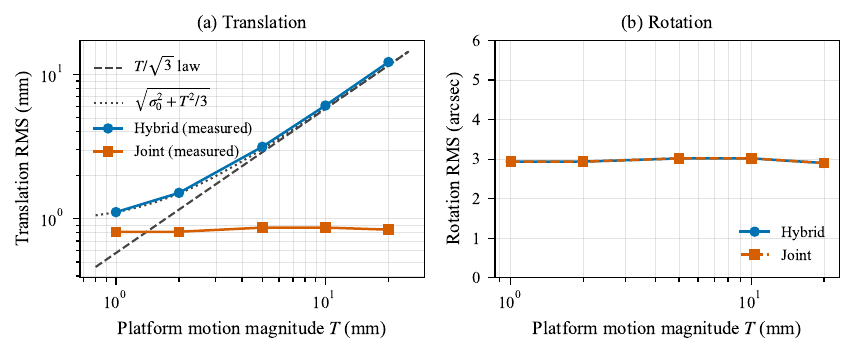}
\caption{Verification of the depth-prior working boundary (multi-camera
configuration): (a) translation RMS versus platform motion magnitude $T$;
the hybrid scheme follows \eqref{eq:quadrature} and approaches the
$T/\sqrt{3}$ paraxial asymptote, while joint recovery remains at its noise
floor; (b) rotation RMS, which remains nearly constant for both tested
schemes over the translation sweep.}
\label{fig:tsweep}
\end{figure}

To verify the one-point prior-bias law, the platform translation magnitude
is swept over $T=1$--20~mm with uniformly distributed directions under the
typical multi-camera condition. Figure~\ref{fig:tsweep} compares the
prior-constrained hybrid estimator with joint 6-DOF recovery. Because the
control point is close to the optical axis, $|\cos\alpha_c|\approx1$.
The measured translation RMSE of the hybrid estimator follows
\eqref{eq:quadrature} with $\sigma_0=0.95$~mm to within 9\% over the tested
range and approaches the $T/\sqrt{3}$ asymptote. The value
$T=\sqrt{3}\sigma_0\approx1.6$~mm marks the point at which the axial-prior
bias reaches the one-point noise floor. Joint recovery remains at its
0.81--0.87~mm noise floor.

The rotation RMSE remains within $2.9$--$3.0''$ for both schemes over the
translation sweep, consistent with the low leakage observed under the tested
geometry. The one-point $\Delta Z$ RMSE of 0.56--0.57~mm in
Table~\ref{tab:onecp} reflects the imposed optical-axis prior, whereas the
0.69-mm two-point result in Table~\ref{tab:twocp} is obtained from full-rank
two-ray recovery. These results are consistent with the prior-bias and
rotation-leakage analyses in Section~\ref{sec:bounds}.

\subsection{Efficiency of the Methods}
\begin{figure}[htbp]
\centering
\includegraphics[width=0.9\columnwidth]{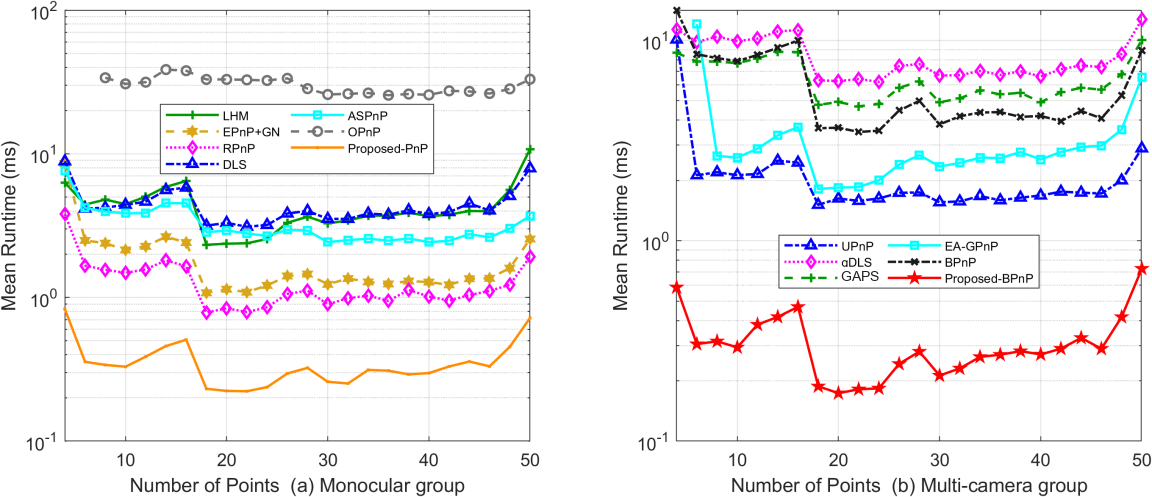}
\caption{Average runtime of all the methods versus the number of measurement
points: (a) monocular group; (b) multi-camera group.}
\label{fig:runtime}
\end{figure}

Fig.~\ref{fig:runtime} reports the average runtime versus the number of
measurement points, and Table~\ref{tab:typical} lists the values under the
typical condition. Comprising two deterministic linear passes, Proposed-BPnP
and Proposed-PnP take only about 0.5~ms per solution. In the monocular
group, Proposed-PnP (0.54~ms) is about $4\times$ faster than the fastest
conventional method RPnP (2.4~ms) and more than an order of magnitude faster
than OPnP; in the multi-camera group, Proposed-BPnP (0.46~ms) is about
$4\times$ faster than the fastest generalized solver UPnP (1.8~ms), roughly
an order of magnitude faster than gDLS/GAPS (7.3--9.7~ms) and the
optimization-based BPnP (9.6~ms), and over $600\times$ faster than gOp
(286~ms), meeting the demand of high-frame-rate real-time
monitoring. Overall, under the practical constraints of $\pm 30'$ attitude
change, $\pm 1$~mm translation, and $\pm 2$~mm measurement-point
perturbation, the re-linearization pass removes the large-rotation
truncation error, and the rigid-body-constrained linear rotation estimation
trades a residual bias of about $1.5''$ from the unmodeled translation for
the best accuracy (typical RMSE $2.97''$ for the multi-camera
configuration), complete freedom from divergence, one sixth of the noise
sensitivity of absolute-pose differencing, and a runtime advantage ranging
from $4\times$ to over $600\times$.

\section{Experiments With Real Data}
\label{sec:real}
\begin{figure}[t]
\centering
\includegraphics[width=0.9\columnwidth]{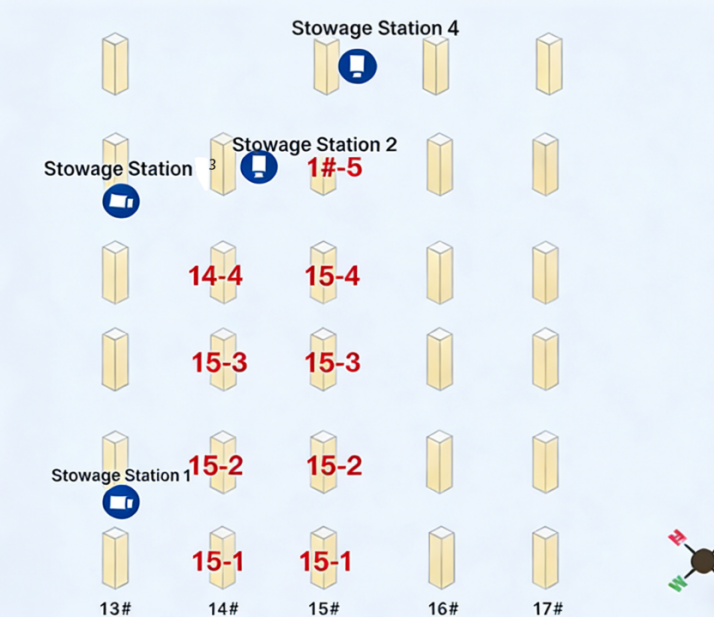}\\[2pt]
\includegraphics[width=0.29\columnwidth]{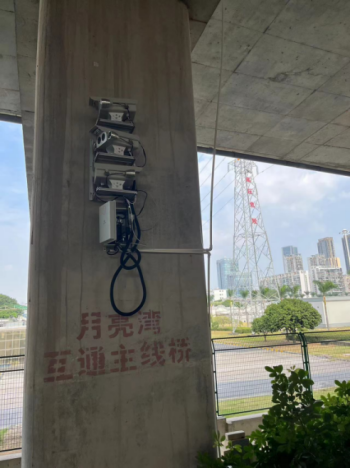}\hspace{0.02\columnwidth}
\includegraphics[width=0.59\columnwidth]{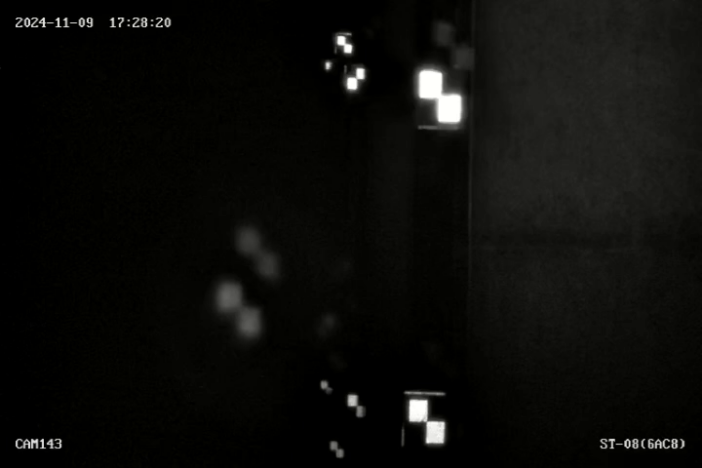}
\caption{Experimental layout: (a) layout diagram (top); (b) on-site
installation (bottom left) and an image captured by one of the station
cameras (bottom right).}
\label{fig:layout}
\end{figure}

We evaluated the control-free rotation mode through deformation measurements
at the Tianyuan Interchange of the Shenzhen Outer-Ring Expressway.
Measurement points were installed on a pier affected by nearby construction,
while no stable control point was available within any camera field of view.
The proposed framework estimated platform rotation directly from the
measurement-point observations and compensated the resulting virtual
displacement. The total-station measurements provide an independent reference
for evaluating the compensated structural displacement. Because translation
is unobservable in this mode, its residual contribution is included in the
reported measurement error. Figure~\ref{fig:layout} shows the experimental
setup.

Four monitoring stations were installed on a pier adjacent to the monitored
pier. Each station contained three cameras and observed 6--10 measurement
points. The initial 3D coordinates of all target points were measured using
a Leica TS60 total station located near the vision measurement platform.
After leveling, the $Z$ axis of the total-station frame was aligned with the
bridge axis, the $Y$ axis pointed vertically downward, and the $X$ axis was
defined according to the right-hand rule.

\subsection{Camera Calibration}
Because no dedicated stable control points were available, the
initial-epoch 3D coordinates of the target points served as calibration
references. These coordinates were measured using the total station, and
the corresponding camera images were acquired synchronously. The cameras
used for deformation measurement were calibrated from the 3D point
coordinates in the total-station frame and their corresponding image
coordinates. The reprojection errors of Cameras 0 and 1 were 0.50 and
0.57 pixels, respectively. These values are consistent with the
approximately 2-pixel/mm object-space sampling rate.
Table~\ref{tab:intrinsics} lists the intrinsic parameters of all three
cameras at Station~1 using their manufacturer serial numbers.

\begin{table}[t]
\centering
\caption{Intrinsic Parameters of the Three Cameras of Station~1 (by
Manufacturer Serial Number)}
\label{tab:intrinsics}
\begin{tabular}{lcccc}
\toprule
Camera & $f_x$ (pixel) & $f_y$ (pixel) & $c_x$ (pixel) & $c_y$ (pixel)\\
\midrule
Camera 122 & 71034.47 & 71034.47 & 1920.5 & 1080.5\\
Camera 123 & 34719.93 & 34719.93 & 1920.5 & 1080.5\\
Camera 125 & 69386.41 & 69386.41 & 1920.5 & 1080.5\\
\bottomrule
\end{tabular}
\end{table}

\subsection{Deformation Measurement}
\begin{figure}[t]
\centering
\includegraphics[width=0.9\columnwidth]{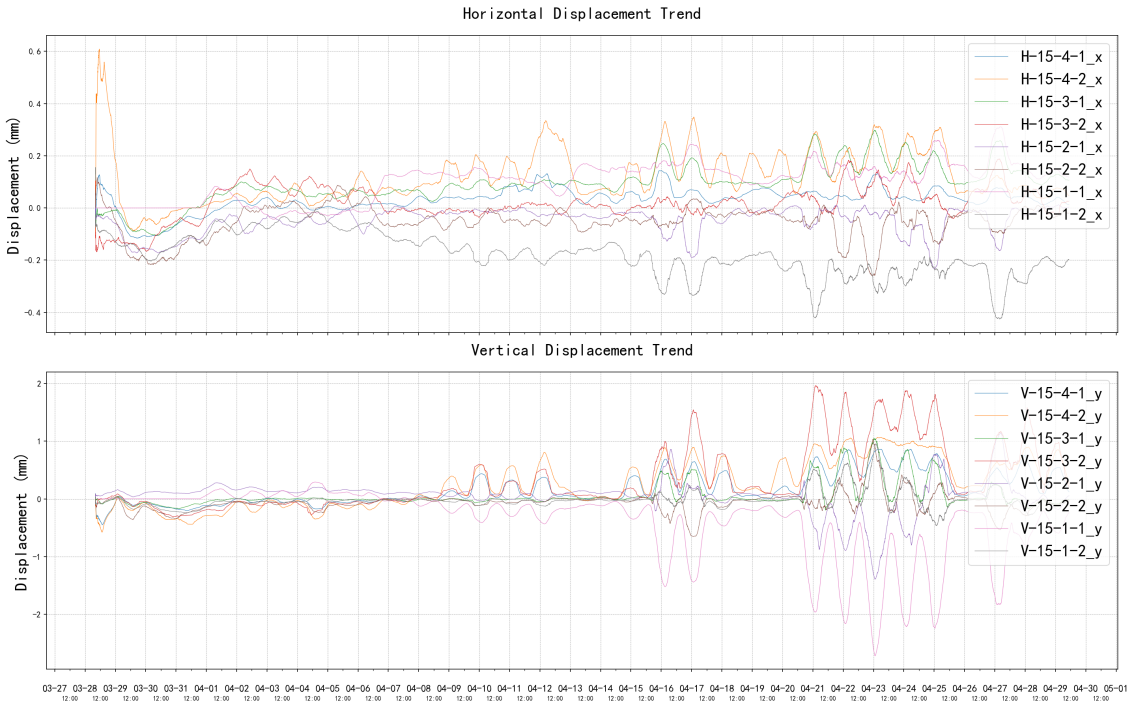}
\caption{Key point deformation measurement results: (a) horizontal
displacement along the $X$-axis; (b) vertical displacement along the
$Y$-axis.}
\label{fig:deform}
\end{figure}

Figure~\ref{fig:deform} presents the recovered horizontal and vertical
displacements of the pier measurement points. Despite the absence of a
stable control field, the proposed method agrees closely with the independent
total-station measurements. Across the 14 coordinate-wise comparisons in
Table~\ref{tab:totalstation}, the median RMS error is 0.85~mm, and 10 of the
14 RMS errors do not exceed 1.1~mm. The complete RMS range is
0.51--2.09~mm.

\begin{table}[t]
\centering
\caption{Comparison of the Measured Displacements With the Total Station}
\label{tab:totalstation}
\setlength{\tabcolsep}{4pt}
\begin{tabular}{lcccc}
\toprule
Point & \makecell*[c]{Mean error\\$X$ (mm)} & \makecell*[c]{RMSE\\$X$ (mm)}
       & \makecell*[c]{Mean error\\$Y$ (mm)} & \makecell*[c]{RMSE\\$Y$ (mm)}\\
\midrule
15-1-1 & 0.58  & 0.70 & 0.06  & 0.71\\
15-1-2 & $-$0.23 & 0.51 & 0.72  & 1.07\\
15-2-1 & $-$0.31 & 0.88 & $-$1.08 & 1.88\\
15-2-2 & $-$0.41 & 0.72 & $-$0.54 & 0.89\\
15-3-1 & 0.53  & 0.81 & $-$0.51 & 0.74\\
15-3-2 & 0.88  & 1.28 & 1.87  & 2.09\\
15-4-1 & 0.31  & 0.66 & $-$1.42 & 1.58\\
\bottomrule
\end{tabular}
\end{table}

\subsection{Robustness to 3D Coordinate Perturbations on Public Sequences}
\label{sec:tum}

The preceding experiments use coded targets and surveyed coordinates. We
further evaluate the method on public sequences containing noncooperative
features. Two TUM RGB-D sequences are used~\cite{tum}. The
\texttt{fr1\_xyz} sequence contains 798 handheld frames and has a scene
depth of approximately 1.2~m. The \texttt{fr2\_desk} sequence contains
2965 frames of a texture-rich desk scene. For each sequence, 100 frame
pairs are selected at frame skips of 1, 5, 10, and 30 to cover small to
moderate inter-frame motions.

Correspondences are obtained using KLT tracking with forward and backward
verification~\cite{lucas1981,shi1994}. The 3D coordinates are
back-projected from the first-frame depth map. All methods receive the same
tracks and 3D coordinates. The baselines independently estimate the
absolute pose of each frame and then compose the relative pose. They include
EPnP~\cite{epnp}, SQPnP~\cite{sqpnp}, an iterative LM solver, and RANSAC
using EPnP hypotheses, a 2-pixel inlier threshold, and 200 iterations.

A trial is classified as divergent when the rotation error exceeds
$5^\circ$ or when the solver fails. The reported RMSE values are computed
over the nondivergent trials. Runtime is measured using single-threaded
Python and OpenCV on a consumer desktop CPU. The motion-capture ground truth
provides millimeter-level translation accuracy and approximately
100-arcsec rotation accuracy. It therefore supports method ranking and
divergence analysis but is not used to validate arcsecond-level accuracy,
which is evaluated by the controlled experiments above.

\begin{table}[t]
\centering
\caption{Rotation RMSE and divergence rates on the clean TUM sequences.
The RMSE is reported in arcseconds, and the divergence rate is reported in
parentheses as a percentage.}
\label{tab:tumbase}
\setlength{\tabcolsep}{2.0pt}
\begin{tabular}{@{}lrrrr@{}}
\toprule
& \multicolumn{2}{c}{\texttt{fr1\_xyz}} &
  \multicolumn{2}{c}{\texttt{fr2\_desk}}\\
Method & skip 1 & skip 30 & skip 1 & skip 30\\
\midrule
Differential (GN) & 1240 (0) & 5747 (2) & 850 (0) & 5730 (9)\\
Differential (linear) & 1240 (0) & 5665 (2) & 850 (0) & 5736 (9)\\
EPnP & 1397 (0) & 5494 (5) & 875 (0) & 6167 (8)\\
SQPnP & 1194 (0) & 5632 (2) & 848 (0) & 5810 (9)\\
Iterative & 1240 (0) & 4997 (9) & 850 (0) & 5754 (10)\\
RANSAC (2\,px) & 1486 (0) & 5036 (0) & 872 (0) & 4431 (9)\\
\bottomrule
\end{tabular}
\end{table}

\begin{table}[t]
\centering
\caption{Robustness to Per-Point 3D Coordinate Perturbations at a Frame
Skip of 1. Rotation RMSE Is Reported in Arcseconds, and the Divergence Rate
Is Reported in Percent.}
\label{tab:tumperturb}
\small
\setlength{\tabcolsep}{3.5pt}
\begin{tabular}{lcccc}
\toprule
Method & clean & $+$20\,mm & $+$50\,mm &
\makecell*[c]{8 pts\\$+$20\,mm}\\
\midrule
Differential (GN) & 1240 (0) & \textbf{1237 (0)} & \textbf{1219 (0)}
 & \textbf{1590 (0)}\\
SQPnP & 1194 (0) & 1193 (0) & 1192 (0) & 1596 (0)\\
EPnP & 1397 (0) & 1400 (0) & 1426 (0) & 2216 (1)\\
Iterative & 1240 (0) & 1858 (13) & 8195 (67) & 2776 (28)\\
RANSAC (2\,px) & 1486 (0) & 7457 (16) & 5626 (86) & 3435 (96)\\
\bottomrule
\end{tabular}
\end{table}

Tables~\ref{tab:tumbase} and~\ref{tab:tumperturb} support three findings.
On clean data, the differential solver matches the strongest PnP baselines
within the ground-truth noise and has an equal or lower divergence rate. The
one-pass result remains within 1.4\% on \texttt{fr1\_xyz} and 0.1\% on
\texttt{fr2\_desk} of the fully iterated result up to $1.6^\circ$
inter-frame motion.

Under shared 3D coordinate perturbations of up to 50~mm, the differential
rotation RMSE remains within $1219''$--$1240''$ on \texttt{fr1\_xyz} and
$845''$--$848''$ on \texttt{fr2\_desk}, with zero divergence. In contrast,
the iterative and RANSAC-based pipelines reach divergence rates of 67\% and
96\%, respectively. The EuRoC \texttt{V1\_01\_easy} stereo
sequence~\cite{euroc} shows the same trend. The differential estimators
remain at $258''$--$259''$ with zero divergence, whereas the 2-pixel RANSAC
result reaches $7820''$ with 13\% divergence.

Correspondence outliers form a separate error channel. With SIFT
matching~\cite{lowe2004}, RANSAC outperforms the nonrobust differential
estimator on clean 3D data, whereas shared coordinate perturbations favor
the differential residual. A robust wrapper around the differential
residual could therefore combine both advantages.

\section{Discussion of Learning-Based Alternatives}
\label{sec:dl}

Learned detectors, matchers, and optical-flow estimators
\cite{detone2018,sarlin2020,sun2021,teed2020} could provide upstream
pixel-displacement observations in noncooperative scenes. However, the
present application requires nearly unbiased localization at the
$10^{-2}$-pixel level and predictable epoch-to-epoch errors. Learned
operators may exhibit texture- and viewpoint-dependent biases of
$10^{-1}$--1 pixel, while equivalent closed-form uncertainty guarantees are
generally unavailable. End-to-end pose regression also operates at much
coarser scales in the cited formulation~\cite{kendall2015}. We therefore
view learning-based components as complementary front ends rather than
replacements for the differential geometric estimator.

\section{Conclusion}
This study presented a differential linear framework that estimates platform
motion and structural displacement directly from inter-frame image
differences. Platform rotation is recovered under an explicit rank condition,
and the effects of unmodeled translation and nonrigid point motion are
quantified by a leakage bound. Translation is prior-constrained with one
surveyed control point and fully observable from two nonparallel control
rays. The analysis also characterizes the total-depth approximation and
conditional re-linearization scales, proves exact exclusion of contamination
confined to control data outside the rotation stage, and derives the exact
$T/\sqrt{3}$ RMS law for the omitted optical-axis component. The framework
inherits exact cancellation of translational extrinsic errors and bounded
sensitivity to rotational extrinsic errors from the underlying differential
model.

Under rotations of up to $\pm30$~arcmin, unmodeled translations of up to
$\pm1$~mm, and 3D point perturbations of up to $\pm2$~mm, the multi-camera
estimator achieved a rotation RMSE of $2.97''$, a translation RMSE of
1.19~mm with one surveyed control point, zero divergence, and an average
runtime of 0.5~ms. It was four times faster than the fastest evaluated
generalized solver and 20 times faster than the optimization-based binocular
method, while remaining robust to the tested extrinsic errors of up to
20~arcmin and 50~mm. On the evaluated public RGB-D and stereo sequences, it
matched the strongest PnP baselines on clean data and maintained zero
divergence under 3D coordinate perturbations that produced divergence rates
of 67\% and 96\% for the iterative and RANSAC-based pipelines. These results
demonstrate state-of-the-art accuracy, robustness, and efficiency among the
evaluated methods while reducing the requirement for dedicated control
points. Future work will address larger inter-frame motions and learned front
ends compatible with differential estimation.

\bibliographystyle{IEEEtran}
\bibliography{ref}

\end{document}